\documentclass[10pt,journal,compsoc]{./IEEEtran}

\ifCLASSOPTIONcompsoc
  \usepackage[nocompress]{cite}
\else
  \usepackage{cite}
\fi

\newcommand{\tablestyle}[2]{\setlength{\tabcolsep}{#1}\renewcommand{\arraystretch}{#2}\centering\footnotesize}
\newlength\savewidth
\usepackage[table]{xcolor}

\ifCLASSINFOpdf
  \usepackage[pdftex]{graphicx}
  \graphicspath{{./pics/}}
  \DeclareGraphicsExtensions{.pdf,.jpeg,.png}
\else
\fi

\usepackage{amsmath}
\usepackage{algorithmic}
\usepackage[linesnumbered,ruled]{algorithm2e}

\usepackage{subcaption}

\usepackage{url}

\usepackage{epsfig}
\usepackage{amssymb}
\usepackage[pagebackref=true,breaklinks=true,letterpaper=true,colorlinks,bookmarks=false]{hyperref}

\usepackage{xcolor}
\usepackage{xspace}
\usepackage{multirow}
\usepackage{booktabs}
\usepackage{placeins}
\usepackage{amssymb}
\usepackage{pifont}
\usepackage{bm}
\usepackage{threeparttable}
\usepackage{makecell}
\usepackage{verbatim}

\usepackage{amsmath,amsfonts,bm}

\def\eqref#1{equation~\ref{#1}}

\def\1{\bm{1}}

\def\vx{{\bm{x}}}
\def\vy{{\bm{y}}}

\DeclareMathAlphabet{\mathsfit}{\encodingdefault}{\sfdefault}{m}{sl}
\SetMathAlphabet{\mathsfit}{bold}{\encodingdefault}{\sfdefault}{bx}{n}

\def\gT{{\mathcal{T}}}

\usepackage{dblfloatfix} 
\usepackage{kantlipsum} 
\usepackage{ragged2e}

\begin{document}

\title{Channel Exchanging Networks for Multimodal and Multitask Dense Image Prediction}

\author{Yikai Wang, Fuchun Sun, Wenbing Huang, Fengxiang He, Dacheng Tao\thanks{
Y. Wang and F. Sun are with  Beijing National Research Center for Infor-mation Science and Technology (BNRist), State Key Lab on Intelligent Technology and Systems, Department of Computer Science and Technology, Tsinghua University.  
W. Huang is with Gaoling School of Artificial Intelligence, Renmin University of China and Beijing Key Laboratory of Big Data Management and Analysis Methods, Beijing, China. F. He and D. Tao are with JD Explore Academy, JD.com Inc.}}

\markboth{IEEE Transactions on Pattern Analysis and Machine Intelligence}
{Shell \MakeLowercase{\textit{et al.}}: Bare Demo of IEEEtran.cls for Computer Society Journals}

\IEEEtitleabstractindextext{
\begin{abstract}
Multimodal fusion and multitask learning are two vital topics in machine learning. Despite the fruitful progress, existing methods for both problems are still brittle to the same challenge---it remains dilemmatic to integrate the common information across modalities (resp. tasks)  meanwhile preserving the specific patterns of each modality (resp. task). Besides,  while they are actually closely related to each other, multimodal fusion and multitask learning are rarely explored within the same methodological framework before. In this paper, we   propose Channel-Exchanging-Network (CEN) which is self-adaptive, parameter-free, and more importantly, applicable for  multimodal  and multitask dense image prediction. At its core, CEN adaptively exchanges channels between subnetworks of different modalities. Specifically, the channel exchanging process is self-guided by individual channel importance that is measured by the magnitude of Batch-Normalization (BN) scaling factor during training. For the application of dense image prediction, the validity of CEN is tested by four different scenarios: multimodal fusion, cycle multimodal fusion, multitask learning, and multimodal multitask learning. Extensive experiments on semantic segmentation via RGB-D data and image translation through multi-domain input verify the effectiveness of CEN compared to state-of-the-art methods. Detailed ablation studies have also been carried out, which  demonstrate the advantage of each component we propose. Our code is available at \url{https://github.com/yikaiw/CEN}.
\end{abstract}

}

\maketitle

\IEEEdisplaynontitleabstractindextext

\IEEEpeerreviewmaketitle

\IEEEraisesectionheading{\section{Introduction}\label{sec:introduction}}

\IEEEPARstart{E}{ncouraged} by the growing availability of low-cost sensors, \emph{multimodal fusion} that takes advantage of multiple data sources for classification or regression  becomes one of the central problems in machine learning~\cite{journals/pami/BaltrusaitisAM19}. Joining the success of deep learning, multimodal fusion is recently specified as \emph{deep multimodal fusion} by introducing end-to-end neural integration of multiple modalities~\cite{ramachandram2017deep}, and it has exhibited remarkable benefits 
against the unimodal paradigm in semantic segmentation~\cite{lin2019refinenet,journals/ijcv/ValadaMB20}, action recognition~\cite{fan2018end,conf/eccv/GarciaMM18,journals/tip/SongLLG20}, visual question answering~\cite{conf/iccv/AntolALMBZP15,conf/nips/IlievskiF17}, and many others~\cite{conf/iccv/BalntasDSSKK17,conf/iclr/JinYBJ19,conf/iccv/ZhangZSWSL19}. \emph{Multitask learning}~\cite{zhang2021survey} is another crucial topic in machine learning. It aims to seek models to solve multiple tasks simultaneously, which enjoys the benefit of model generation and data efficiency against the methods that learn each task independently. Similar to multimodal fusion, multitask learning has also been developed from previously shallow methods~\cite{zhou2017multi} to deep variants~\cite{misra2016cross,liu2019end,guo2020learning,standley2020tasks,sun2020adashare} by taking advantage of deep learning. The successful applications of multitask learning include navigation~\cite{andreas2017modular}, robot manipulation~\cite{rahmatizadeh2018vision}, etc.

In general, dense image prediction could be a collection of computer vision tasks that aim at classifying (\emph{e.g.}, segmentation~\cite{journals/corr/LongSD14,lin2019refinenet,DBLP:journals/pami/ChenPKMY18,DBLP:conf/iccv/HuangLC021}) or regressing (\emph{e.g.}, image-to-image translation~\cite{conf/cvpr/IsolaZZE17,DBLP:conf/eccv/HuangLBK18,DBLP:journals/inffus/MaML19,DBLP:journals/ijcv/LeeTMHLSY20}) every pixel in an image, namely, producing pixel-wise output based on the given input pixels.  The learning pipeline for dense prediction is usually expected to capture  rich spatial details or strong semantics, which  also benefits greatly  from multimodal data sources or the multitask joint training. A variety of works tailored for dense image prediction have been done towards multimodal fusion and multitask learning. For multimodal fusion, regarding the type of how they fuse, existing methods are generally categorized into \emph{aggregation-based} fusion~\cite{conf/accv/HazirbasMDC16,conf/cvpr/ZengTHYSCW19,journals/ijcv/ValadaMB20}, \emph{alignment-based} fusion~\cite{journals/tip/SongLLG20,conf/eccv/WangWTSW16}, and the mixture of them~\cite{journals/pami/BaltrusaitisAM19}.  As for multitask learning, in the context of deep learning, two types of contemporary techniques are identified: \emph{hard parameter-sharing}~\cite{kokkinos2017ubernet,chennupati2019multinet++} and \emph{soft parameter-sharing}~\cite{misra2016cross,ruder2019latent}.  
Despite the fruitful progress, existing methods for both problems are still brittle to the same challenge---it remains dilemmatic to integrate the common information across modalities (resp. tasks)  meanwhile preserving the specific patterns of each modality (resp. task) for multimodal fusion (resp. multitask learning). To be more specific, for multimodal fusion, the aggregation-based fusion is prone to underestimating the intra-modal propagation, whereas the alignment-based fusion mostly delivers ineffective inter-modal fusion owing to the weak message exchanging by solely training  alignment losses~\cite{conf/cvpr/DuWWZW19,conf/iccv/LeePH17,conf/cvpr/ZengTHYSCW19}. A similar issue exists in multitask learning. Current hard/soft parameter sharing schemes could be vulnerable to the negative transfer issue across different tasks owing to the insufficient balance between inter-task knowledge sharing and intra-task information processing~\cite{sener2018multi}. When focusing on dense image prediction,  multimodal fusion and multitask learning can also be regarded as the dual problem of each other. As will be described in \textsection~\ref{sec:method}, multimodal fusion corresponds to the multiple-input-single-output problem while multitask learning, inversely, is of the single-input-multiple-output formulation. Yet, most previous works study these two problems separately without revealing their common property.

In this paper, we propose Channel-Exchanging-Network (CEN) which is self-adaptive, parameter-free, and applicable for multimodal  and multitask dense image prediction. For unification, we refer to both the modality-specific network in multimodal fusion and the task-specific network in multitask learning as a subnetwork. To enable message passing among different modalities/tasks, CEN adaptively exchanges the channels between subnetworks. The core of CEN lies in its smaller-norm-less-informative assumption inspired by network pruning~\cite{conf/iccv/LiuLSHYZ17,conf/iclr/YeL0W18}. To be specific, we utilize the scaling factor (\emph{i.e.}, $\gamma$) of Batch-Normalization (BN)~\cite{conf/icml/IoffeS15} or Instance-Normalization (IN)~\cite{DBLP:journals/corr/UlyanovVL16} as the importance measurement for each corresponding channel, and replace the channels associated with close-to-zero factors of each subnetwork with the mean of other subnetworks. Such message exchanging is self-adaptive in determining when to exchange, and hence it is capable of accomplishing better trade-off between inter-subnetwork knowledge sharing and intra-subnetwork information processing, in contrast to conventional multimodal and multitask learning methods. Further, the channel exchanging operation itself is parameter-free, making CEN less prone to overfitting, while, for example, the attention-based fusion~\cite{journals/ijcv/ValadaMB20} needs extra parameters to adjust the importance of each subnetwork. Another hallmark of CEN is that the encoder parameters except for BN layers of all subnetworks are shared with each other (\textsection~\ref{sec:mm}). Apart from compacting the model size, we apply the idea  here to serve specific purposes in CEN: by using private BNs, we can determine the channel importance for each individual modality;
by sharing convolutional filters, the corresponding channels among different modalities are embedded with the same mapping, thus more capable of modelling the modality-common statistic. 

CEN is generally powerful, capable of addressing four different problems in image dense prediction: multimodal fusion, cycle multimodal fusion, multitask learning, and multimodal multitask learning. For multimodal fusion, we conduct channel exchanging on the encoder side to allow information integration between different input modalities. We also design cycle multimodal fusion to reuse the knowledge among different generation flows, which can promote performance for each flow. As natural extensions, channel exchanging could be applied to the decoder side or both the decoder and encoder to exchange task-specific information for multitask learning or for multimodal multitask learning. These details will be provided in \textsection~\ref{sec:method}.

To sum up, our contributions are as follows:
\begin{itemize}
    \item We propose CEN for message fusion, which is self-adaptive and parameter-free. The core of CEN is to replace the channels associated with close-to-zero BN or IN scaling factors of each subnetwork with the mean of others. 
    \item CEN is generally powerful and is applied to multimodal fusion, cycle multimodal fusion, multitask learning, and multimodal multitask learning. To the best of our knowledge, it is the first time that one single technique is explicitly employed to address multimodal fusion, multitask learning, or both, particularly on dense image prediction. 
    \item Experimental evaluations are conducted on semantic segmentation via RGB-D data~\cite{conf/eccv/SilbermanHKF12,conf/cvpr/SongLX15} and image translation through multi-domain input~\cite{conf/cvpr/ZamirSSGMS18}. It demonstrates that CEN yields remarkably superior performance to various kinds of multimodal fusion methods and multitask learning methods under a fair condition of comparison. 
\end{itemize}

\section{Related Work} \label{sec:related}
We introduce the methods of deep multimodal fusion and deep multitask learning, especially using  dense image prediction as examples. We also discuss other related concepts.

\textbf{Deep multimodal fusion.}
Regarding dense image prediction, deep multimodal fusion uses multiple data sources to enhance pixel-level semantics and fine-grained details against the single-modality counterpart. To this end, related methods toward dense image prediction are basically categorized into aggregation-based fusion and alignment-based fusion.  Aggregation-based fusion methods apply a certain operation (\emph{e.g.}, averaging~\cite{conf/accv/HazirbasMDC16}, concatenation~\cite{conf/icml/NgiamKKNLN11,conf/cvpr/ZengTHYSCW19}, and attention-based modules~\cite{journals/ijcv/ValadaMB20,DBLP:journals/expert/ZhouYLL21}) to fuse high-resolution feature maps and combine multimodal subnetworks into a single network. For example, U2Fusion~\cite{DBLP:journals/pami/XuMJGL22} concatenates source images and puts forward  the information measurement for unsupervised learning. RDFNet \cite{conf/iccv/LeePH17} adopts multi-layer fusion and iteratively refines  fused features with additional convolutional blocks for aggregation.  Due to the weakness in intra-modal processing, recent aggregation-based works perform feature fusion while still maintaining the subnetworks of all modalities~\cite{conf/cvpr/DuWWZW19,conf/iccv/LinCCHH17}. Alignment-based fusion methods~\cite{journals/tip/SongLLG20,conf/eccv/WangWTSW16}, instead, adopt regulation losses to align the embedding of  subnetworks while keeping full propagation for each of them. These methods align multimodal features by applying the similarity regulation, where Maximum-Mean-Discrepancy (MMD)~\cite{journals/jmlr/GrettonBRSS12} is usually adopted for the measurement.
However, simply focusing on unifying the whole distribution may overlook the specific patterns in each domain/modality \cite{conf/nips/BousmalisTSKE16,journals/tip/SongLLG20}. Hence, \cite{conf/eccv/WangWTSW16} provides a way that might alleviate this issue, which correlates modality-common features while simultaneously maintaining modality-specific information. Another categorization of multimodal fusion towards dense prediction could be generally specified as early, middle, and late fusion, depending on when to fuse, which have been discussed in earlier works~\cite{atrey2010multimodal,bruni2014multimodal,hall1997introduction,snoek2005early} and also in the current deep learning literature~\cite{journals/pami/BaltrusaitisAM19,lazaridou2014wampimuk,wang2020asymfusion,conf/nips/VriesSMLPC17}. Besides, evaluations in \cite{conf/iccv/LeePH17} indicate that the single-layer fusion   can not effectively exploit multimodal features, especially for addressing high-resolution predictions torward dense image prediction. \cite{conf/accv/HazirbasMDC16} points out that the performance of dense feature fusion is highly affected by the choice of which layer to fuse. Beyond dense image prediction,
 there are  other portions of the multimodal learning literature, \emph{e.g.}, based on modulation~\cite{de2017guesswhat,dumoulin2018feature,conf/nips/VriesSMLPC17}. Different from these categories of fusion methods, we propose a new fusion method by channel exchanging, which potentially enjoys the guarantee of both sufficient inter-model interactions and intra-modal learning. 

\textbf{Deep multitask learning.} In general, multitask visual perception predicts multiple output domains based on one same vision domain. Typical approaches could include  designing hard parameter-sharing and soft parameter-sharing. Specifically, hard parameter-sharing imposes a fixed subset of hidden layers to be shared across  tasks and other layers to be task-specific, for example, UberNet~\cite{kokkinos2017ubernet}, U2Fusion~\cite{DBLP:journals/pami/XuMJGL22}, and  others~\cite{long2015learning,chennupati2019multinet++,suteu2019regularizing}.  Differently, for soft (or partial) parameter-sharing, there could be a separate set (or a significant fraction) of parameters per task, and  models are correlated either by adaptive feature sharing or by aligning parameters to be similar, for example,  Cross-stitch~\cite{misra2016cross}, Sluice~\cite{ruder2019latent}, and NDDR~\cite{gao2019nddr}. Yet, compared with the learning upon single modalities, multitask learning is not always beneficial, since the performance might be harmed by the negative transfer (negative knowledge transfer across tasks), which is discussed in~\cite{DBLP:journals/corr/ZhangY17aa,DBLP:conf/icml/StandleyZCGMS20,DBLP:conf/cvpr/ZamirSCSCMG20}. In addition, many multitask learning methods are  specifically designed for dense image prediction, which is also the main focus of this paper. For example, MTI-Net~\cite{DBLP:conf/eccv/VandenhendeGG20} distills dense features across different tasks with multimodal feature aggregation. ~\cite{DBLP:conf/iccv/ZhuPIE17, DBLP:conf/cvpr/ZamirSCSCMG20} explicitly enforce cycle-based consistency between domains to improve performance and generalization. U2Fusion~\cite{DBLP:journals/pami/XuMJGL22} develops joint training and sequential training that leverages a shared model to handle multitask learning for image-to-image translation. In this paper, we integrate the benefits of both hard parameter-sharing and soft parameter-sharing. Specifically, for multitask learning, we share the parameters of encoders for all tasks (hard parameter-sharing) and then conduct CEN on decoders (soft parameter-sharing).

\textbf{Other related concepts.} 
The idea of using the BN scaling factor to evaluate the importance of CNN channels has been studied in network pruning~\cite{conf/iccv/LiuLSHYZ17,conf/iclr/YeL0W18} and representation learning~\cite{shao2020channel}. 
Moreover, \cite{conf/iccv/LiuLSHYZ17} enforces $\ell_1$ norm penalty on the scaling factors and explicitly prunes out filters meeting sparsity criteria. Here, we apply this idea as an adaptive tool to determine where to exchange and fuse.  CBN~\cite{conf/nips/VriesSMLPC17} performs cross-modal message passing by modulating BN of one modality conditional on the other, which is different from our method that directly exchanges channels across modalities for fusion. 
ShuffleNet~\cite{conf/cvpr/ZhangZLS18} proposes to shuffle a portion of channels among multiple groups for efficient propagation in light-weight networks, which is similar to our idea of exchanging channels for message fusion. Yet, while the motivation of our paper is highly different, the exchanging process is self-determined by the BN scaling factors, instead of the random exchanging in ShuffleNet.

\section{Channel Exchanging Networks}
\label{sec:method}
We first introduce the general formulation of CEN, and then follow it up by specifying the design of four different settings: multimodal fusion, cycle multimodal fusion, multitask learning, and multimodal multitask learning. 

\subsection{The general mechanism}
\label{subsec:channel_exc}

For either multimodal or multitask learning, we are interested in studying the relationship between subnetworks on different streams of input-output pairs. Suppose we have the data of $M$ streams $\{(\vx_{m}, \vy_{m})\}_{m=1}^M$, where $\vx_m$ and $\vy_m$ represent the input data point and output label, respectively.
The subnetwork of the $m$-th stream is dubbed as $f_m$. 
The notion of ``stream'' can be flexibly specified: for multimodal fusion, a different stream corresponds to a different modality where 
$\vx_m$ varies but $\vy_m$ keeps unchanged in terms of different $m$; for multitask learning, on the contrary, a different  stream implies a different task, where $\vx_m$ usually keeps the same and $\vy_m$ represents the label for task $m$.

A trivial training paradigm is minimizing the loss of each subnetwork $f_m$ independently, which leads to the loss between the prediction $\hat{\bm{y}}_m:=f_m(\bm{x}_m)$ and the label $\bm{y}_m$\footnote{Note that this loss should be summed over all data points in real implementation. Here we consider a single data point throughout the paper for simplicity.},
\begin{equation}
\label{eq:loss}
    \min_{f_{1:M}}\sum_{m=1}^{M}\mathcal{L}\big(f_m(\bm{x}_m),\bm{y}_m\big).
\end{equation}
However, the independent training strategy fails to characterize the affinity between different streams, limiting the expressivity of multimodal information fusion or multitask knowledge transfer.

In this work, we propose CEN that adaptively exchanges the knowledge between different subnetworks in an end-to-end manner. In form, the training objective in Eq.~\ref{eq:loss} can be rewritten as
\begin{equation}
\label{eq:cen}
\min_{f_{1:M}}\sum_{m=1}^{M}\mathcal{L}\big(f_m(\bm{x}_{1:M}),\bm{y}_m\big)+\lambda\|\hat{\bm{\gamma}}_{m}\|_1
\end{equation}
where,
\vspace{-0.3em}
\begin{itemize}
    \item 
    The subnetwork $f_m(\bm{x}_{1:M})$ (instead of $f_m(\bm{x}_m)$ in Eq.~\ref{eq:loss}) fuses multimodal information by channel exchanging from other subnetworks to the $m$-th subnetwork, as we will detail later;
    \item Each subnetwork is equipped with BN layers containing the scaling factors $\bm{\gamma}_{m}$, and we will penalize the $\ell_1$ norm of their certain portion $\bm{\hat{\gamma}}_{m}$ for sparsity. The $\ell_1$ norm is uniformly applied to all BN layers. Here, we omit the layer index for simplicity.
\end{itemize}

Prior to introducing the mechanism of channel exchanging, we first review the Batch-Normalization (BN) layer~\cite{conf/icml/IoffeS15}, which is used widely in deep learning to eliminate covariate shift and improve generalization. For a certain BN layer, we denote by $\bm{x}_{m}$ the feature map of the $m$-th subnetwork, and by $\bm{x}_{m,c}$ the $c$-th channel. The BN layer performs a  normalization of $\bm{x}_{m}$ followed by an affine transformation, namely,
\begin{equation}
\label{eq:bn}
\bm{x}'_{m,c} = \gamma_{m,c}\frac{\bm{x}_{m,c}-\mu_{m,c}}{\sqrt{\sigma^2_{m,c}+\epsilon}}+\beta_{m,c},
\end{equation}
where, $\mu_{m,c}$ and $\sigma_{m,c}$ compute the mean and the standard deviation, respectively, of all activations over all pixel locations ($H$ and $W$) for the current mini-batch data; $\gamma_{m,c}$ and $\beta_{m,c}$ are the trainable scaling factor and offset, respectively; $\epsilon$ is a small constant to avoid divisions by zero. The following layer takes $\{\bm{x}'_{m,c}\}_c$ as input after a non-linear function.

The factor $\gamma_{m,c}$ in Eq.~\ref{eq:bn} evaluates the correlation between the input $\bm{x}_{m,c}$ and the output $\bm{x}'_{m,c}$ during training. The gradient of the loss \emph{w.r.t.} $\bm{x}_{m,c}$ will approach 0 if $\gamma_{m,c}\rightarrow 0$ at one training step, implying that $\bm{x}_{m,c}$ will almost lose its influence to the final prediction and become redundant thereby at this traing step.

In addition, as will be shown in Fig.~\ref{pic:scaling_factor} (a), if the scaling factor of one channel (with  sparsity constraints) is lower than the small threshold at one training step, this channel will hardly recover and almost become redundant during the later training process.

\begin{figure}[t!]
\centering
\hskip-0.03in
\includegraphics[scale=0.257]{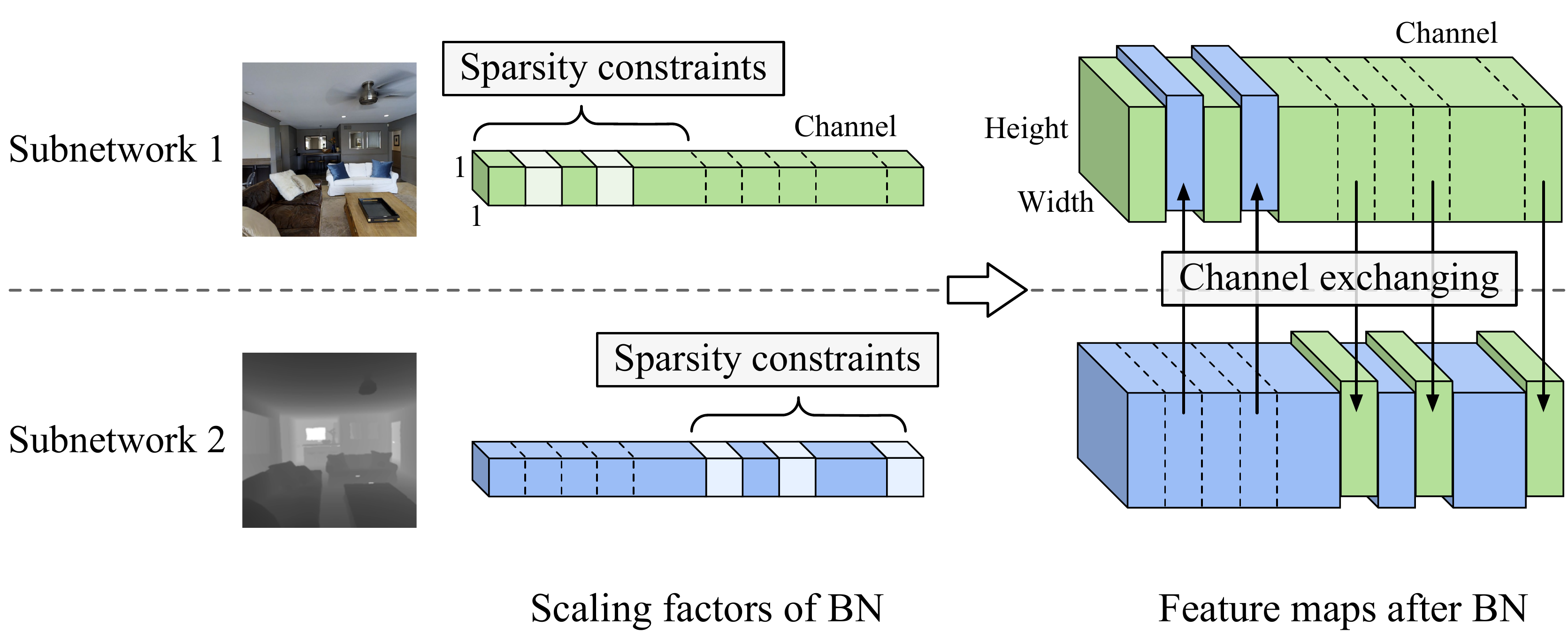}
\vskip -0.05 in
\caption{An illustration of CEN. The sparsity constraints on scaling factors are applied to disjoint channel regions of different modalities. A feature map will be replaced by that of other modalities at the same position, if its scaling factor is lower than a threshold. }
\label{channel}
\vskip -0.1 in
\end{figure}

\begin{figure*}[t!]
\centering
\includegraphics[scale=0.55]{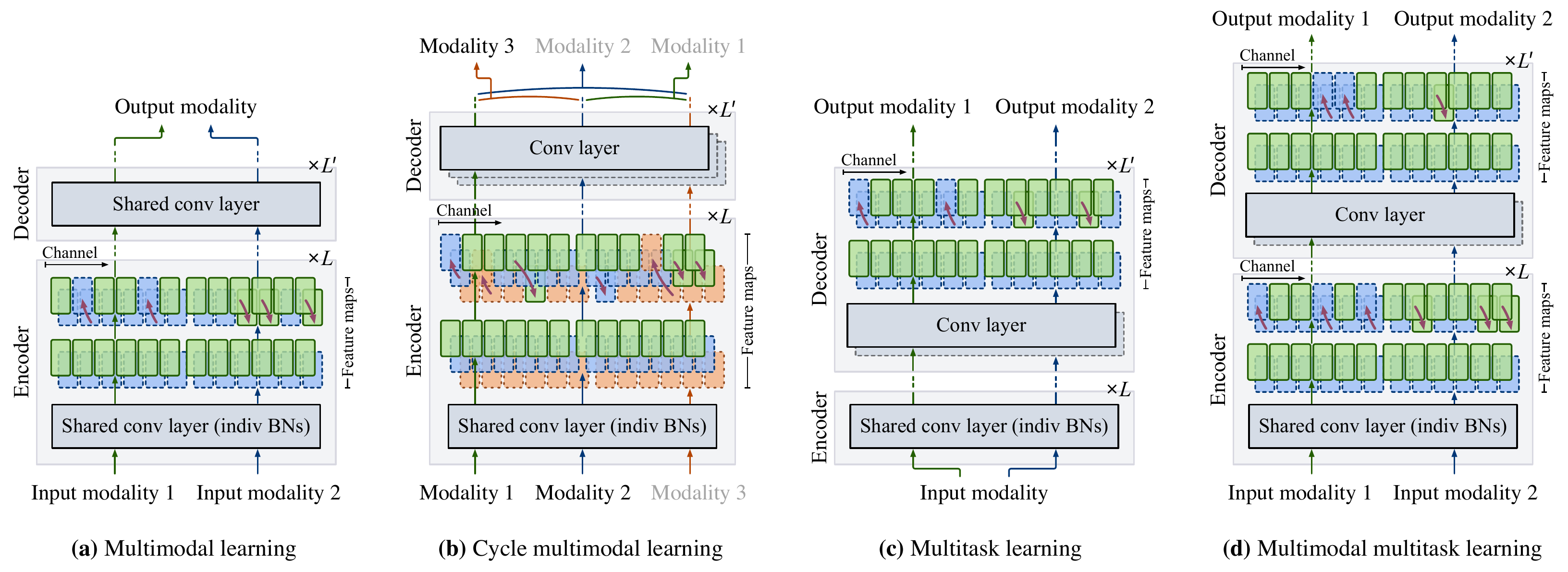}
\caption[]{Structures of CENs for multimodal fusion, cycle multimodal fusion, multitask learning, and multimodal multitask learning.  For cycle multimodal learning, given the case with three modalities, only two of the three forward passes are performed at each time. Here, ``conv'' and ``indiv'' are abbreviations for ``convolutional'' and ``individual'', respectively. $L$ and $L'$ denote  layer numbers of the encoder and the decoder, respectively. }
\vspace{-0.06in}
\label{pic:sketch}
\end{figure*}

It  motivates us to replace the channels of small scaling factors with the ones of other subnetworks, since those channels potentially are redundant. To do so, we  have 
\begin{equation}
\label{eq:exchange-bn}
\hskip-0.2em
\bm{x}'_{m,c} =
\begin{cases}
\gamma_{m,c}\frac{\bm{x}_{m,c}-\mu_{m,c}}{\sqrt{\sigma^2_{m,c}+\epsilon}}+\beta_{m,c}, \quad\quad\;\;\,\text{if}\;\; \gamma_{m,c}>\theta; \\
\frac{1}{M-1}\sum\limits_{m'\neq m}^M\gamma_{m',c}\frac{\bm{x}_{m',c}-\mu_{m',c}}{\sqrt{\sigma^2_{m',c}+\epsilon}}+\beta_{m',c}, \;\;\text{else};
\end{cases}
\end{equation}
where the current channel is replaced with the mean of other channels if its scaling factor is smaller than a certain threshold $\theta\approx0^+$. In a nutshell, if one channel of one modality has little impact on the final prediction, then we replace it with the mean of other modalities. We apply Eq.~\ref{eq:exchange-bn} for each modality before feeding them into the nonlinear activation followed by the convolutions in the next layer. Gradients are detached from the replaced channel and back-propagated through the new ones.
    
Fig.~\ref{channel} illustrates our channel exchanging process in each of the layers. In order to  In our implementation, we equally divide the whole channels into $M$ sub-parts and only perform the channel exchanging in each corresponding sub-part for each modality. This is mainly to avoid a portion of channels being redundant \emph{w.r.t.} all modalities. More detailed reasons are described in \textsection~\ref{sec:discussion}. We denote the scaling factors that are allowed to be replaced as $\bm{\hat{\gamma}}_{m}$. We further impose the sparsity constraint on $\bm{\hat{\gamma}}_{m}$ in Eq.~\ref{eq:cen} to discover unnecessary channels. As the exchanging in Eq.~\ref{eq:exchange-bn} is a directed process within only one sub-part of channels, it hopefully can not only retain modal-specific propagation in the other $M-1$ sub-parts but also avoid unavailing exchanging since $\gamma_{m',c}$, different from $\hat{\gamma}_{m,c}$, is out of the sparsity constraint.

Regarding  specific  tasks  where  Instance-Normalizations (INs) are  used for normalization instead of BNs, the sparsity constraints are similarly applied to scaling factors of INs, and the channel exchanging design (Eq.~\ref{eq:exchange-bn}) is still applicable.

We summarize the advantages of our CEN below:
\begin{itemize}
    \item \textbf{Prameter-free}. As specified in Eq.~\ref{eq:exchange-bn}, CEN involves no additional parameter and applies BN scaling factors to control the exchanging process. 
    \item \textbf{Self-adaptive}. The channel exchanging  could take place at every layer throughout the encoder or/and decoder. BN scaling factors are learned from the data, which adaptively balances the inter-subnetwork processing and inter-subnetwork fusion.
\end{itemize}

\subsection{Multimodal fusion via CEN on encoders}
\label{sec:mm}

In this part, we focus particularly on multimodal fusion $\{\vx_m\}_{m=1}^M\hskip-0.03in\rightarrow\hskip-0.03in\vy$, where   $\vx_m$ denotes the $m$-th input modality, and all subnetworks generate the same output $\vy$, \emph{i.e.}, $\vy_m=\vy,\forall m=1,\cdots,M$. Given that this paper mainly copes with dense prediction problems (such as depth estimation or semantic segmentation), the subnetwork $f_m$ is of the encoder-decoder style. The goal of multimodal fusion is to effectively fuse the information of all modalities to improve the prediction accuracy for the target output. It is thus natural to fix the same decoder for all subnetworks and conduct CEN between their encoders. The architecture of multimodal fusion is depicted in Fig.~\ref{pic:sketch} (a).

We first carry out sparsity penalty on  BN scaling factors for the $m$-th encoder following Eq.~\ref{eq:cen}, and then perform channel exchanging. Besides, the final output of the decoder is an ensemble of all modalities associated with the decision scores $\{\alpha_m\}_{m=1}^M$\footnote{The decision scores are learnable  scalars, optimized by comparing  ensembled outputs with labels while temporally freezing (detaching) the subnetworks. The decision scores are fixed during inference. }; in our implementation, these decision scores are learned by an additional softmax output to meet the simplex constraint $\sum_{m=1}^M\alpha_m=1$.

It is known in~\cite{conf/cvpr/ChangYSKH19} that leveraging individual BN layers characterizes the traits of different domains or modalities. In our method, specifically, different scaling factors (Eq.~\ref{eq:bn}) evaluate the importance of the channels of different modalities, and they should be decoupled.
With the exception of BN (or IN) layers, all subnetworks share all  parameters (\emph{e.g.} convolutional filters\footnote{If the input channels of different modalities are different (\emph{e.g.}, RGB and depth), we will broaden their sizes to be the same as their Least Common Multiple (LCM).}) in the encoder with each other. The hope is that we can further reduce the network complexity and therefore improve the predictive generalization. Rather, considering the specific design of our framework, sharing convolutional filters is able to capture the common patterns in different modalities, which is a crucial purpose of multimodal fusion. This design further compacts the multimodal architecture to be almost as small as the unimodal one, as will be evaluated in Table~\ref{tabs:our_mmplementation}. In our experiments, we conduct multimodal fusion on RGB-D images or on other domains of images corresponding to the same image content. In this scenario, all modalities are homogeneous in the sense that they are just different views of the same input. Thus, sharing parameters between different subnetworks still yields promisingly expressive power. Nevertheless, when we are dealing with heterogeneous modalities (\emph{e.g.}, images and text sequences), it would impede the expressive power of the subnetworks if keeping sharing their parameters, hence a more dexterous mechanism is suggested, and the discussion of which is left for future exploration.

\subsection{Cycle multimodal fusion via CEN on encoders}
\label{sec:cycle_fusion}
In the previous section (\textsection~\ref{sec:mm}), we have introduced how to apply CEN on multimodal fusion. Here, we discuss a more complicated setting: cycle multimodal fusion. Assuming we have $\{\vx_m\}_{m=1}^M\rightarrow\vx_{M+1}$, where the output is specified as the $(M+1)$-th modality for consistent denotation. Note that such learning task is related to a different task $\{\vx_m\}_{m=1,m\neq j}^{M+1}\rightarrow\vx_{j}$, which, inversely, uses modality $M+1$ along with the remaining modalities to generate modality $j$. Actually, we can go through all the $M+1$ cases by cycling different output modality, which leads to a set of cycle multimodal fusion tasks $\{\gT_j:=\{\vx_m\}_{m=1,m\neq j}^{M+1}\rightarrow\vx_{j}\}_{j=1}^{M+1}$.

\vspace{0.02in}
By~\textsection~\ref{sec:mm}, a straightforward way is applying CEN independently to each multimodal fusion task $\gT_j$ for fusing the input modalities. Nevertheless, such an independent learning fashion is unable to reveal the relationships between $\gT_j$s. Although different tasks conduct different generation directions, these tasks are tackling overlapping modalities, hence potentially, their learning knowledge might be  reused and the learning processes could be coupled. Towards this purpose, we enforce all $\gT_j$s to share the same encoder  except the BN parameters. Specifically, for each task $\gT_j$, we utilize distinct sets of BN parameters for different input modalities, giving rise to the total number of BN parameter sets for all tasks as $M(M+1)$. With the separated BNs, we then carry out CEN on the encoder for multimodal fusion for each task $\gT_j$. The sketched pipeline is illustrated in Fig.~\ref{pic:sketch} (b). Note that for the case with three modalities, channels are still divided into two parts, since for cycle multimodal fusion, only two of the three modalities are sent to the encoder at each time.

Obviously, cycle multimodal fusion is a multitask generalization of the multimodal fusion in \textsection~\ref{sec:mm}. The key benefit  is that it simultaneously addresses all combinations of the cycling generation tasks with only one single pair of the encoder and decoder, which dramatically decreases the model complexity. More interestingly, as we will demonstrate in our experiments, the cycle multimodal fusion can improve each of the single-task multimodal fusion,  probably thanks to the knowledge transfer by parameter sharing and joint training. We will provide more details and evaluations for cycle multimodal fusion in the experiment section.

\subsection{Multitask learning via CEN on decoders}
\label{sec:multitask_decoder}
Different from multimodal fusion, multitask learning requires to predict different labels for different subnetworks: $\vx\rightarrow\{\vy_m\}_{m=1}^M$, where we assume all tasks have the same input, \emph{i.e.}, $\vx_m=\vx, \forall m=1,\cdots,M$ and the output label is $\vy_m$ for the task $m$. 
The advantage of multitask learning is to improve model generalization and data efficiency, by sharing task-common knowledge while retaining task-specific information.  One of the widely-used methods is employing the hard parameter-sharing mechanism~\cite{DBLP:conf/cvpr/Kokkinos17} that shares the encoder and uses task-specific decoders. Despite its popularity in previous applications, modelling the multitask relationship by solely sharing the encoder is insufficient in characterizing high-level patterns, particularly the related features across decoders.  

To address the aforementioned issues, we propose to perform channel exchanging on the decoders.  Our goal of employing CEN on decoders lies in adaptively discovering the redundant channels in decoders and compensating for the information from the channels of other tasks. The methodology is illustrated  is in Fig.~\ref{pic:sketch} (c). Specifically, the sparsity penalty of BN (or IN) scaling factors is added to the decoder part. Accordingly, for the $m$-th subnetwork,  channel exchanging is conducted from other decoders to the $m$-th decoder.

\subsection{Multimodal multitask learning via CEN on both encoders and decoders}
It could be straightforward to combine the designs in~\textsection~\ref{sec:mm} and \textsection~\ref{sec:multitask_decoder} to handle multimodal multitask learning tasks, with multiple input and output modalities, as illustrated in Fig.~\ref{pic:sketch} (d). It requires to address $\{\{\vx_{m_1}\}_{m_1=1}^{M_1}\rightarrow\vy_{m_2}\}_{m_2=1}^{M_2}$, where $M_1$ and $M_2$ are the numbers of input and output modalities, respectively. To enable simultaneous  multimodal fusion and multitask learning, we perform CEN on both encoders and decoders. The input for each decoder is given by CEN on all encoders. In this case, we share the convolutional layers at the encoder part and privatize $M_1M_2$ groups of BN (or IN) parameters. Similarly, for the $m_2$-th task/decoder (where $m_2=1,\cdots,M_2$), we adopt  $\{\alpha_{m_1}^{m_2}\}_{m_1=1}^{M_1}$ as  decision scores for ensemble that meet $\sum_{m_1=1}^{M_1}\alpha_{m_1}^{m_2}=1$.

\section{Experiments}
\label{sec:experiments}

We contrast the performance of CEN against existing methods on the four problems in Fig.~\ref{pic:sketch}. For multimodal fusion, we conduct experiments on the two tasks: semantic segmentation and image-to-image translation. For the other three problems, we evaluate the performance mainly on image-to-image translation, since this task contains a rich number of image modalities and is suitable for evaluations under various settings. The datasets and implementation details for semantic segmentation and image-to-image translation are provided below.

\textbf{Semantic segmentation.} 
We evaluate our method on two public datasets NYUDv2~\cite{conf/eccv/SilbermanHKF12} and SUN RGB-D~\cite{conf/cvpr/SongLX15}, which consider RGB and depth as input. Regarding NYUDv2, we follow the standard settings and adopt the split of 795 images for training and 654 for testing,  predicting standard 40 classes~\cite{conf/cvpr/GuptaAM13}. SUN RGB-D is one of the most challenging large-scale benchmarks for indoor semantic segmentation, containing 10,335 RGB-D images of 37 semantic classes. We use the public train-test split (5,285 vs 5,050).
We consider RefineNet~\cite{lin2019refinenet}/PSPNet~\cite{conf/cvpr/ZhaoSQWJ17} as our segmentation framework whose backbone is implemented by ResNet~\cite{conf/cvpr/HeZRS16} pretrained from ImageNet dataset \cite{journals/ijcv/RussakovskyDSKS15}. 
The initial learning rates are set to $5\times10^{-4}$ for the encoder and $3\times10^{-3}$  for the  decoder, respectively, both of which are reduced to their halves every 100/150 epochs (of total epochs 300/450) on NYUDv2 with ResNet101/ResNet152 and every 20 epochs (of total epochs 60) on SUN RGB-D. The mini-batch size, momentum, and weight decay are selected as 6, 0.9, and $10^{-5}$, respectively, on both datasets. 
We set $\lambda=5\times10^{-3}$ in Eq.~\ref{eq:cen} and the threshold to $\theta=2\times10^{-2}$ in Eq.~\ref{eq:exchange-bn}.
Unless otherwise specified, we adopt the multi-scale strategy~\cite{conf/iccv/LeePH17,lin2019refinenet} during the test time. 
We employ common evaluation metrics including Mean IoU, Pixel Accuracy, and Mean Accuracy \cite{lin2019refinenet}. Full implementation details are provided in the appendix.

\textbf{Image-to-image translation.} 
We adopt Taskonomy~\cite{conf/cvpr/ZamirSSGMS18}, a dataset with 4 million images of indoor scenes gathered from about 600 buildings. Each image in Taskonomy has more than 10 multimodal representations, including depth (euclidean/zbuffer), shade, normal, texture, edge, principal curvature, etc. For efficiency, we sample 1,000 high-quality multimodal images for training, and 500 for validation. We also provide experiments with 15,000 sampled images for training in the appendix.
Following Pix2pix~\cite{conf/cvpr/IsolaZZE17}, we adopt the U-Net-256 structure for image translation with the consistent setups with~\cite{conf/cvpr/IsolaZZE17}. The BN computations are replaced with Instance Normalization layers (INs), and our method (Eq.~\ref{eq:exchange-bn}) is still applicable. We adopt individual INs in the encoder, and share all other parameters including INs in the decoder. We set $\lambda$ to $10^{-3}$ for sparsity constraints and the threshold $\theta$ to $10^{-2}$. FID~\cite{conf/nips/HeuselRUNH17} and KID~\cite{conf/iclr/BinkowskiSAG18} are adopted as evaluation metrics, as will be introduced in the appendix.

\vspace{-0.05in}
\subsection{Evaluations on multimodal fusion}
We first assess the importance of each component in CEN solely on the semantic segmentation dataset NYUDv2, and then compare the performance with other multimodal fusion baselines and SOTA methods on semantic segmentation and image-to-image translation.

\begin{table}[t]
\centering
\caption{Detailed results for different versions of our CEN on NYUDv2. All results are obtained with the backbone RefineNet (ResNet101) of single-scale evaluation for test. ``Ens.'' is the  abbreviation for ``Ensemble''.}
\tablestyle{3.5pt}{1.1}
\resizebox{0.98\linewidth}{!}{
\begin{tabular}{lllc|ccc}
\toprule[1pt]
\multirow{2}*{Convs}&\multirow{2}*{\makecell[l]{BNs}}&\multirow{2}*{$\ell_1$ Regulation}&\multirow{2}*{Exchange}&\multicolumn{3}{c}{Mean IoU (\%)}\\
&&&&RGB & Depth &Ens.$\,$\\
\midrule
Unshared&Unshared&$\;\;\;\;\;\;\;\;\times$&$\times$&45.5&35.8&47.6\\
Shared & Shared&$\;\;\;\;\;\;\;\;\times$&$\times$&{43.7}&{35.5}&{45.2}\\
Shared&Unshared&$\;\;\;\;\;\;\;\;\times$&$\times$&46.2&38.4&48.0\\
\midrule
Shared&Unshared&$\;\;\;\;\;\;\;\;\times$&$\checkmark$ (fixed 30\%)&44.9&40.3&47.2\\
Shared&Unshared&$\;\;\;\;\;\;\;\;\times$&$\checkmark$ (random)$\;\;\,$&44.2&40.5&46.8\\
\midrule
Unshared &Unshared &All-channel&$\times$&44.6&35.3&46.6\\
Unshared &Unshared &All-channel&$\checkmark$&46.8&41.7&49.1\\
Shared &Unshared &All-channel&$\times$&46.1&37.9&47.5\\
Shared &Unshared &All-channel&$\checkmark$&48.6&{39.0}&49.8\\
\midrule
Unshared &Unshared  & Half-channel&$\times$&45.1&35.5&47.3\\
Unshared &Unshared  & Half-channel&$\checkmark$&{46.5}&{41.6}&{48.5}\\
Shared &Unshared  & Half-channel&$\times$&46.0&38.1&47.7\\
Shared &Unshared  & Half-channel &$\checkmark$&\textbf{49.7}&\textbf{45.1}&\textbf{51.1}\\
\bottomrule[1pt]
\end{tabular}}
\label{tabs:component}
\end{table}

\begin{figure*}[t]
\centering
\hskip -0.1 in
\includegraphics[scale=0.25]{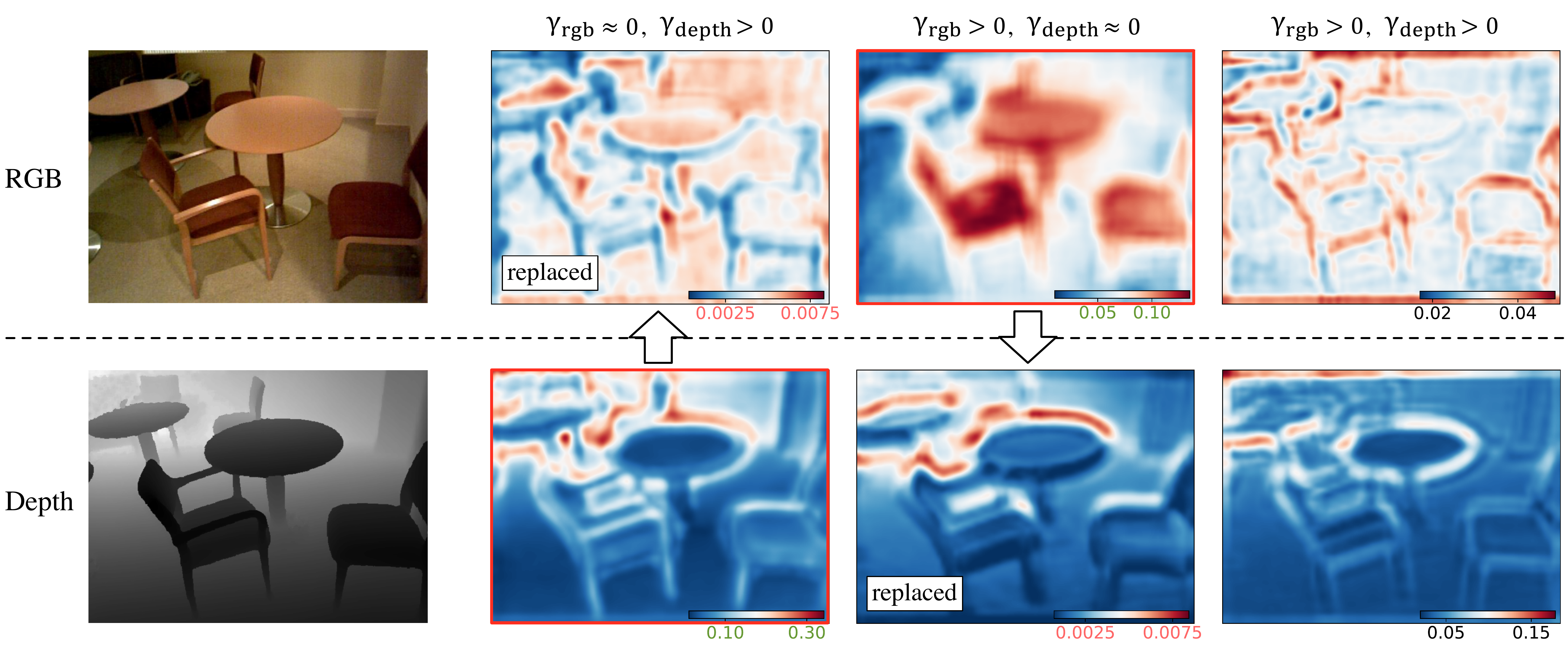}
\caption{ Visualization of the \emph{averaged} feature maps for RGB and Depth. From left to right: the input images, the channels of $(\gamma_{rgb}\approx 0, \gamma_{depth}> 0)$, $(\gamma_{rgb}> 0, \gamma_{depth}\approx 0)$, and $(\gamma_{rgb}> 0, \gamma_{depth}> 0)$. The feature maps are collected in a single layer, specifically, the 9th layer of ResNet, \emph{i.e.} the 2nd layer of the 3rd stage (with 256 channels) of ResNet.  Values under color bars correspond to the  actual values of averaged feature maps.}
\label{featuremap5}
\end{figure*}

\begin{table*}[t]
\centering
\caption{Comparison with three typical fusion methods including concatenation (concat), fusion by alignment (align), and self-attention (self-att.) on NYUDv2. All results are obtained with RefineNet (ResNet101) of single-scale evaluation for test. ``Ens.'' is the  abbreviation for ``Ensemble''.}
\tablestyle{6pt}{1.1}
\resizebox{155mm}{!}{
\begin{tabular}{ll|cc|cc|c}
\toprule[1pt]
\multirow{3}*{Modality}&\multirow{3}*{Approach}&\multicolumn{2}{c|}{Commonly-used setting}&\multicolumn{2}{c|}{Same with our setting}&\multirow{3}*{\makecell[l]{Params. used\\for fusion (M)}}\\
&&{\makecell[c]{Mean IoU (\%)}} &{$\;\;$ \makecell[l]{Params.\\in total (M)}} &{\makecell[c]{Mean IoU (\%)\\RGB / Depth / Ens.}} &{$\;$ \makecell[l]{Params.\\in total (M)}} & \\
\midrule
 RGB & Uni-modal&45.5&118.1&45.5 / $\;\;\,$-$\;\;\,$ / $\;\;$-$\;\;\;$&118.1&-\\
 Depth & Uni-modal&35.8&118.1&$\;$-$\;\;$ / 35.8 / $\;\;$-$\;$&118.1&-\\
\midrule
\multirow{15}{*}{RGB-D}
& Concat (early)&47.2&120.1&47.0 / 37.5 / 47.6&118.8&0.6\\
& Concat (middle)&46.7&147.7&46.6 / 37.0 / 47.4&120.3&2.1\\
& Concat (late)&46.3&169.0&46.3 / 37.2 / 46.9&126.6&8.4\\
& Concat (all-stage)&47.5&171.7&47.8 / 36.9 / 48.3&129.4&11.2\\
\cmidrule(r){2-7}
& Align (early)&46.4&238.8&46.3 / 35.8 / 46.7&120.8&2.6\\
& Align (middle)&47.9&246.7&47.7 / 36.0 / 48.1&128.7&10.5\\
& Align (late)&47.6&278.1&47.3 / 35.4 / 47.6&160.1&41.9\\
& Align (all-stage)&46.8&291.9&46.6 / 35.5 / 47.0&173.9&55.7\\
\cmidrule(r){2-7}
& Self-att. (early)&47.8&124.9&47.7 / 38.3 / 48.2&123.6&5.4\\
& Self-att. (middle)&48.3&166.9&48.0 / 38.1 / 48.7&139.4&21.2\\
& Self-att. (late)&47.5&245.5&47.6 / 38.1 / 48.3&203.2&84.9\\
& Self-att. (all-stage)&48.7&272.3&48.5 / 37.7 / 49.1&231.0&112.8\\
\cmidrule(r){2-7}
& Our CEN&-&-&\textbf{49.7} / \textbf{45.1} / \textbf{51.1}&\textbf{118.2}&\textbf{0.0}\\
\bottomrule[1pt]
\end{tabular}}
\label{tabs:our_mmplementation}
\end{table*}

\begin{figure*}[h]
\centering
\includegraphics[scale=0.41]{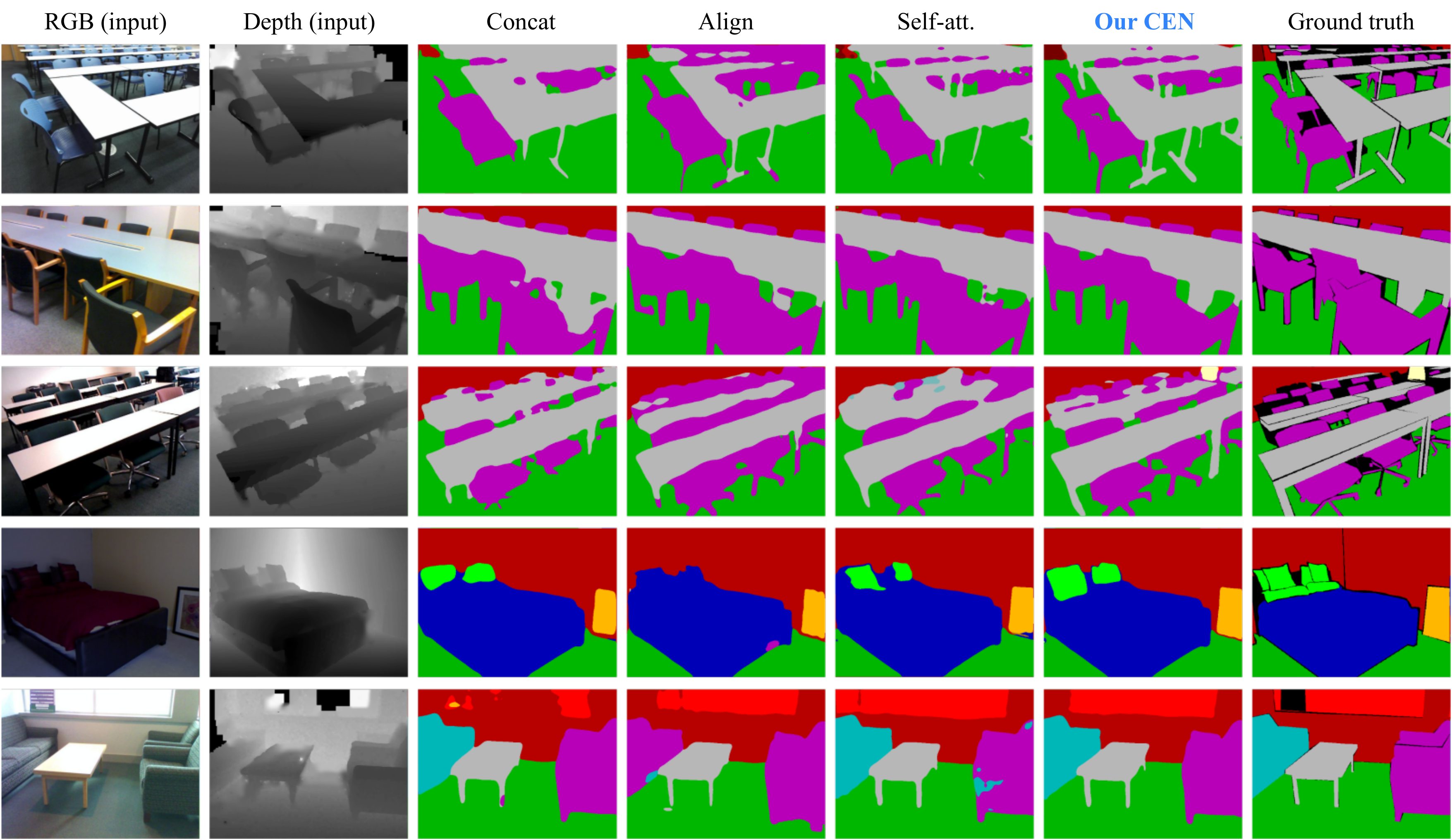}
\caption{Visualization results of semantic segmentation. Images are collected from NYUDv2 and SUN RGB-D datasets. All results are obtained with the backbone RefineNet (ResNet101) of single-scale evaluation for test. }
\label{pic:seg_results}
\end{figure*}

\begin{table*}[t]
\centering
\caption{Comparison with SOTA semantic segmentation  methods on NYUDv2 and SUN RGB-D datasets. $\dag$ indicates our implemented results. Evaluation metrics include Pixel Accuracy, Mean Accuracy and Mean IoU.}
\tablestyle{4pt}{1.1}
\resizebox{155mm}{!}{
\begin{tabular}{lll|ccc|ccc}
\toprule[1pt]
\multirow{3}*{Modality}&\multirow{3}*{Approach}&\multirow{3}*{\makecell[l]{Backbone \\Network}}&\multicolumn{3}{c|}{NYUDv2}&\multicolumn{3}{c}{SUN RGB-D}\\
&&&\makecell[l]{Pixel Acc.\\(\%)} & \makecell[l]{Mean Acc.\\(\%)} & \makecell[l]{Mean IoU\\(\%)}& \makecell[l]{Pixel Acc.\\(\%)} & \makecell[l]{Mean Acc.\\(\%)} & \makecell[l]{Mean IoU\\(\%)} \\
\midrule
\multirow{3}*{RGB} 
&FCN-32s  \cite{journals/corr/LongSD14}&VGG16&60.0&42.2&29.2&68.4&41.1&29.0\\
&RefineNet \cite{lin2019refinenet}&ResNet101&73.8&58.8&46.4&80.8&57.3&46.3\\
&RefineNet \cite{lin2019refinenet}&ResNet152&74.4&59.6&47.6&81.1&57.7&47.0\\
\midrule
\multirow{15}*{RGB-D} 
& FuseNet  \cite{conf/accv/HazirbasMDC16}&VGG16 &68.1&50.4&37.9&76.3&48.3&37.3\\
&ACNet \cite{conf/icip/HuYFW19}&ResNet50&-&-&48.3&-&-&48.1\\
&SSMA \cite{journals/ijcv/ValadaMB20}&ResNet50&75.2&60.5&48.7&81.0&58.1&45.7\\
&SSMA \cite{journals/ijcv/ValadaMB20} $\dag$&ResNet101&75.8&62.3&49.6&81.6&60.4&47.9\\
&CBN~\cite{conf/nips/VriesSMLPC17} $\dag$ & ResNet101&75.5&61.2&48.9&81.5&59.8&47.4\\
&3DGNN \cite{conf/iccv/QiLJFU17}&ResNet101&-&-&-&-&57.0&45.9\\
&SCN \cite{journals/tcyb/LinZJLH20} &ResNet152&-&-&49.6&-&-&50.7\\
& CFN \cite{conf/iccv/LinCCHH17} & ResNet152&-&-&47.7&-&-&48.1\\
&RDFNet \cite{conf/iccv/LeePH17} &ResNet101&75.6&62.2&49.1&80.9&59.6&47.2\\
&RDFNet \cite{conf/iccv/LeePH17} &ResNet152&76.0&62.8&50.1&81.5&60.1&47.7\\
\cmidrule(r){2-9}
&Ours-RefineNet (single-scale) &ResNet101&76.2&62.8&51.1&82.0&60.9&49.6\\
&Ours-RefineNet
&ResNet101&{77.2}&{63.7}&{51.7}&{82.8}&{61.9}&{50.2}\\
&Ours-RefineNet (single-scale) &ResNet152&77.0&64.4&51.6&82.3&61.7&50.0\\
&Ours-RefineNet &ResNet152&{77.4}&{64.8}&{52.2}&{83.2}&{62.5}&{50.8}\\
&Ours-PSPNet &ResNet152&\textbf{77.7}&\textbf{65.0}&\textbf{52.5}&\textbf{83.5}&\textbf{63.2}&\textbf{51.1}\\
\bottomrule[1pt]
\end{tabular}}
\label{tabs:seg_results}
\vskip -0.1 in
\end{table*}

\begin{figure*}[h]
\centering
\includegraphics[scale=0.14]{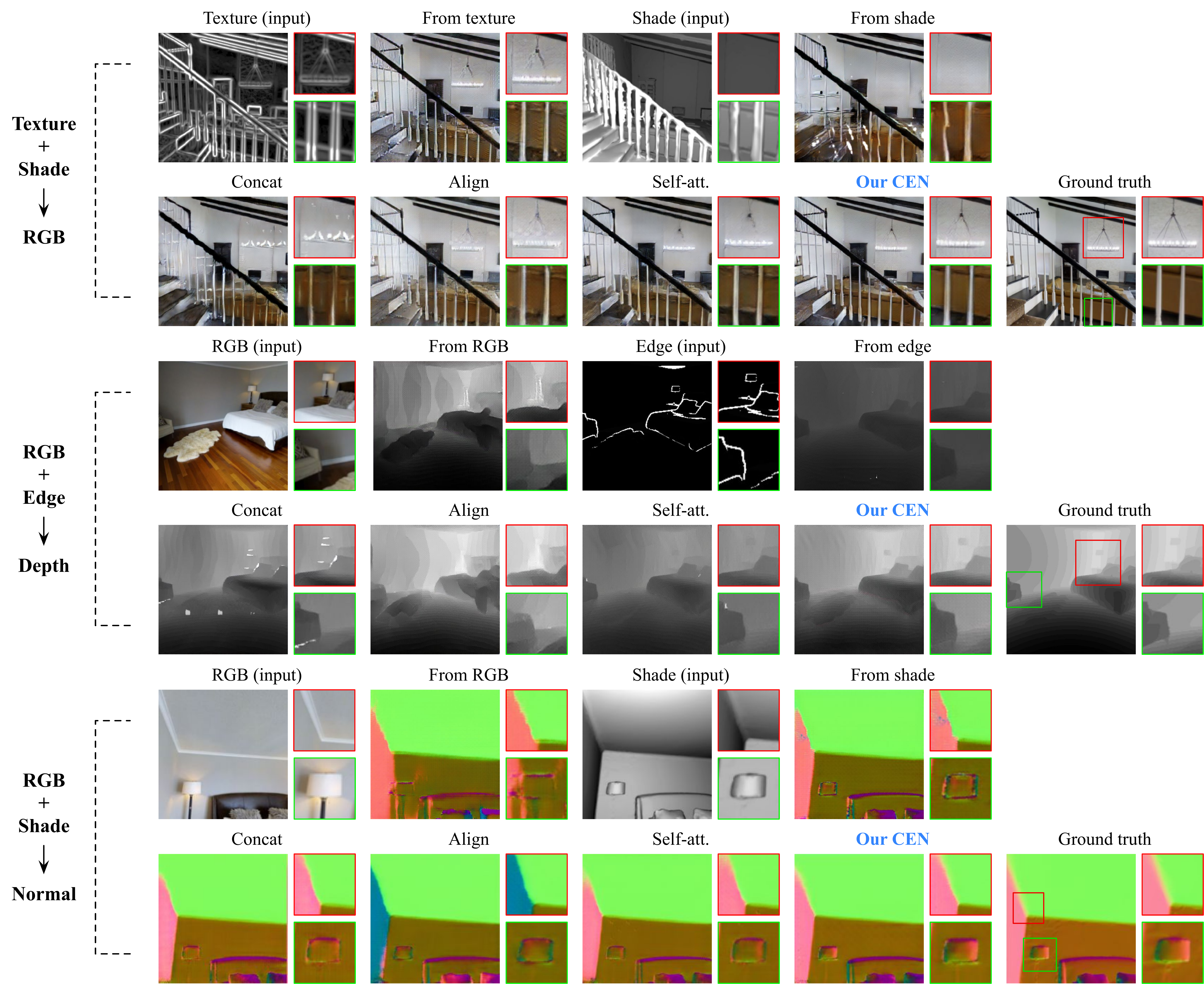}
\caption{Visualization results of multimodal image-to-image translation including Texture+Shade$\to$RGB (top group), RGB+ Edge$\to$Depth (middle group), and RGB+Shade$\to$Normal (bottom group), respectively. The resolution of each predicted image is $256\times256$. More visualizations are provided in the appendix.}
\vspace{-0.5em}
\label{pic:trans_1}
\end{figure*}

\subsubsection{Semantic Segmentation}

\begin{table*}[t]
\centering
\vspace{0.2em}
\caption{Comparison on multimodal image-to-image translation task. Evaluation metrics are FID/KID ($\times 10^{-2}$) for RGB predictions and MAE ($\times 10^{-2}$)/MSE ($\times 10^{-2}$) for other predictions. Lower values indicate better performance for all metrics.}
\vspace{-0.2em}
\tablestyle{8pt}{0.9}
\resizebox{150mm}{!}{
\begin{tabular}{c|c|lccccc}
\toprule[1pt]
$\;$Modality&Our CEN&Baseline&Early&Middle&Late&All-layer\\
\midrule
\multirow{4}{*}{\makecell[l]{Shade+Texture\\\;\;\;$\to$RGB}} 
&\multirow{4}{*}{\textbf{62.63} / \textbf{1.65}}\;\;
& Concat&{87.46 / 3.64}$\;\,$&{95.16 / 4.67}$\;\,$&{122.47 / 6.56}$\;\;\;\,$&{78.82 / 3.13}\;\;\\
&& Average&{93.72 / 4.22}$\;\,$&{93.91 / 4.27}$\;\,$&{126.74 / 7.10}$\;\;\;\,$&{80.64 / 3.24}\;\;\\
&&Align&{99.68 / 4.93}$\;\,$&{95.52 / 4.75}$\;\,$&{\;\;98.33 / 4.70}$\;\;\;\,$&{92.30 / 4.20}\;\;\\
&&Self-att.&{83.60 / 3.38}$\;\,$&{90.79 / 3.92}$\;\,$&{105.62 / 5.42}$\;\;\;\,$&{73.87 / 2.46}\;\;\\
\midrule
\multirow{4}{*}{\makecell[l]{Depth+Normal\\\;\;\,\;$\to$RGB}} 
&\multirow{4}{*}{\textbf{84.33} / \textbf{2.70}}\;\;
& Concat&{105.17 / 5.15}\;\;\;\,&{100.29 / 3.37}\;\;\;\,&{116.51 / 5.74}\;\;\;\,&{\;\;99.08 / 4.28}\;\;\;\,\\
&& Average&{109.25 / 5.50}\;\;\;\,&{104.95 / 4.98}\;\;\;\,&{122.42 / 6.76}\;\;\;\,&{\;\;99.63 / 4.41}\;\;\;\,\\
&&Align&{111.65 / 5.53}\;\;\;\,&{108.92 / 5.26}\;\;\;\,&{105.85 / 4.98}\;\;\;\,&{105.03 / 4.91}\;\;\;\,\\
&&Self-att.&{100.70 / 4.47}\;\;\;\,&{\;\;98.63 / 4.35}\;\;\;\,&{108.02 / 5.09}\;\;\;\,&{\;\;96.73 / 3.95}\;\;\;\,\\
\midrule
\multirow{4}{*}{\makecell[l]{RGB+Shade\\\;$\to$Normal}} 
&\multirow{4}{*}{\makecell{\textbf{11.23} / \textbf{25.09}}}& Concat&{13.34 / 28.27}&{12.15 / 26.54}&{13.93 / 28.80}&{13.36 / 28.51}\\
&& Average&{14.24 / 30.47}&{12.62 / 27.02}&{14.01 / 28.95}&{12.82 / 28.28}\\
&&Align&{14.50 / 31.07}&{13.92 / 29.34}&{12.81 / 27.55}&{15.18 / 32.50}\\
&&Self-att.&{12.99 / 28.21}&{11.75 / 25.86}&{14.22 / 29.07}&{12.63 / 27.61}\\
\midrule
\multirow{4}{*}{\makecell[l]{RGB+Normal\\\;\;$\to$Shade}} 
&\multirow{4}{*}{\makecell{\textbf{11.03} / \textbf{17.16}}}& Concat&{15.62 / 24.49}&{13.81 / 21.24}&{12.62 / 19.17}&{12.83 / 20.18}\\
&& Average&{14.63 / 22.88}&{12.83 / 20.42}&{15.11 / 23.92}&{12.28 / 18.64}\\
&&Align&{13.88 / 22.62}&{13.16 / 21.55}&{12.73 / 20.41}&{14.09 / 22.05}\\
&&Self-att.&{12.14 / 18.26}&{11.52 / 17.33}&{14.47 / 22.82}&{11.79 / 17.62}\\
\midrule
\multirow{4}{*}{\makecell[l]{RGB+Edge\\$\to$Depth}} 
&\multirow{4}{*}{\makecell{\textbf{2.75} / \textbf{6.60}}}& Concat&{3.43 / 7.53}&{3.17 / 7.39}&{3.82 / 7.87}&{3.25 / 7.46}\\
&& Average&{3.62 / 7.78}&{3.41 / 7.64}&{3.56 / 7.73}&{3.30 / 7.44}\\
&&Align&{4.38 / 8.93}&{3.86 / 8.16}&{4.19 / 8.61}&{4.38 / 9.03}\\
&&Self-att.&{3.03 / 7.05}&{3.32 / 7.29}&{3.40 / 7.47}&{3.01 / 6.98}\\
\bottomrule[1pt]
\end{tabular}}
\label{tabs:translation}
\vskip -0.03 in
\end{table*}

\begin{table*}[t!]
\centering
\caption{Multimodal fusion on image translation (to RGB) with   $1\sim4$ input modalities.}
\vspace{-0.3em}
\tablestyle{6pt}{1.05}
\resizebox{150mm}{!}{
\begin{tabular}{c|cccc|ccc}
\toprule[1pt]
Modality&Depth&Normal&Texture&Shade&\makecell[l]{Depth+Normal}&\makecell[l]{Depth+Normal\\+Texture}&\makecell[l]{Depth+Normal\\+Texture+Shade}\\
\midrule
FID&113.91 &108.20 &97.51 &100.96   &84.33&60.90&57.19\\
KID ($\times 10^{-2}$)& 5.68&5.42&4.82&5.17 & 2.70& 1.56&1.33\\
\bottomrule[1pt]
\end{tabular}}
\label{tabs:multimodal}
\end{table*}

\textbf{The validity of each proposed component.}
Table \ref{tabs:component} summarizes the results of different variants of CEN on NYUDv2. We have the following observations:
\begin{itemize}
    \item Compared to the unshared baseline, sharing  convolutional parameters greatly boosts the performance, particularly on the Depth modality (35.8 vs 38.4). Yet, the performance will encounter a clear drop if we additionally share the BN layers. This observation is consistent with our analyses in~\textsection~\ref{sec:mm} due to the different roles of convolutional filters and BN parameters.
    \item As $\ell_1$ enables the discovery of unnecessary channels, naively exchanging channels with a fixed portion (without using $\ell_1$) could not reach good performance. For example, exchanging a fixed portion of 30\% channels (close to the averaged number of exchanged channels in CEN) only gets IoU 47.2. Besides, we try to exchange channels randomly like ShuffleNet or directly discard unimportant channels without channel exchanging, the IoUs of which are 46.8 and 47.5, respectively. 
    \item After carrying out directed channel exchanging under the $\ell_1$ regulation, our model gains a huge improvement on both modalities, \emph{i.e.} from 46.0 to 49.7 on RGB, and from 38.1 to 45.1 on Depth, and finally increases the ensemble Mean IoU from 47.6 to 51.1. It thus verifies the effectiveness of our proposed mechanism on this task.
    \item Note that the channel exchanging is only available on a certain portion of each layer, \emph{i.e.}, exchanging only half of the channels in the two-modal case. When we remove this constraint and allow all channels to be exchanged by Eq.~\ref{eq:exchange-bn}, the accuracy decreases, which we conjecture is owing to the detriment by impeding modal-specific propagation, if all channels are engaged in cross-modal fusion. 
\end{itemize}

After training CEN (with sparsity constraints on disjoint channel regions, as illustrated in Fig.~\ref{channel}), each certain channel belongs to one of the three  categories: $(\gamma_{rgb}\approx 0, \gamma_{depth}> 0)$, $(\gamma_{rgb}> 0, \gamma_{depth}\approx 0)$, and $(\gamma_{rgb}> 0, \gamma_{depth}> 0)$. There will not be $(\gamma_{rgb}\approx 0, \gamma_{depth}\approx 0)$ since we apply sparsity constraints on disjoint channels. 
To further explain why channel exchanging works, Fig. \ref{featuremap5} displays the averaged feature maps of RGB and Depth. Here, ``averaged'' means: Firstly, extracting feature maps at all specific channels (in a layer) that belong to the same (aforementioned) category; Secondly, averaging these feature maps along the channels. We observe from Fig. \ref{featuremap5} that  RGB channels with non-zero scaling factors mainly characterize the texture, while  Depth channels with non-zero factors focus more on the boundary; in this sense, performing channel exchanging can better combine the complementary properties of both modalities.

\textbf{Comparison with  fusion baselines.}
In Table~\ref{tabs:our_mmplementation}, we report  comparison results of our CEN with two aggregation-based methods: concatenation~\cite{conf/cvpr/ZengTHYSCW19} and self-attention~\cite{journals/ijcv/ValadaMB20}, and one alignment-based approach~\cite{conf/eccv/WangWTSW16},  using the same backbone. All baselines are implemented with the early, middle, late, and all stage fusion. For a more fair comparison, all baselines are further  conducted under the same setting (except channel exchanging) with ours, namely, sharing convolutions with individual BNs, and preserving the propagation of all subnetworks (with also the ensemble). Full details are provided in the appendix. It demonstrates that, in both settings, our method always outperforms others by an average improvement of larger than 2\%. We also report the parameters used for fusion, \emph{e.g.} the aggregation weights of two modalities in concatenation. While self-attention (all-stage) attains the closest performance to ours (49.1 vs 51.1), its parameters used for fusion are considerable, whereas our fusion is parameter-free. 

Visualizations are provided in Fig. \ref{pic:seg_results}. We choose the hard cases including the images containing tables and chairs, as well as those with low/high light intensity. We observe that the concatenation method is more sensitive to noises in the depth input. Both concatenation and self-attention methods are weak in predicting thin objects, \emph{e.g.}, table legs and chair legs. These objects are usually missed in the depth input, which may hinder the prediction results after fusion. On the contrary, the prediction results of our method preserve more details and are more robust to the light intensity.

\textbf{Comparison with SOTAs.}
In Table~\ref{tabs:seg_results}, we  contrast our method against a wide range of state-of-the-art methods. Their results are directly copied from previous papers if provided or re-implemented by us otherwise, as marked with annotations. 
Results conclude that our method equipped with PSPNet (ResNet152) achieves new records remarkably superior to previous methods in terms of all metrics on both datasets. In particular, given the same backbone, our method is still much better than RDFNet~\cite{conf/iccv/LeePH17}. To isolate the contribution of RefineNet in our method, Table~\ref{tabs:seg_results} also provides the uni-modal results, where we observe a clear advantage of multimodal fusion.

\begin{table*}[t]
\centering
\caption{Experimental results of cycle multimodal fusion. Evaluation metrics are FID/KID ($\times 10^{-2}$) for RGB predictions and MAE ($\times 10^{-2}$)/MSE ($\times 10^{-2}$) for other predictions. Lower values indicate better performance for all these metrics. ``Curve'' and ``SemSeg'' are abbreviations for the principle curve and semantic segmentation, respectively.}
\tablestyle{4pt}{0.97}
\resizebox{150mm}{!}{
\begin{tabular}{c|p{2.5cm}<{\centering}p{2.65cm}<{\centering}p{2.5cm}<{\centering}p{2.5cm}<{\centering}}
\toprule[1pt]
$\;$Modality&\makecell[c]{CEN (IN$\times$6, \\enc$\times$3, dec$\times$3)}&\makecell[c]{CEN-random (IN$\times$6, \\enc$\times$1, dec$\times$3)} &\makecell[c]{CEN-cycle (IN$\times$6, \\enc$\times$1, dec$\times$3)}&\makecell[c]{CEN-cycle (IN$\times$6, \\enc$\times$1, dec$\times$1)}\\
\midrule
RGB+Shade$\;\to\;$Texture& 1.74 / 3.05 & 2.17 / 4.53& \textbf{1.54} / \textbf{2.56}& {1.62} / {2.81}\\
RGB+Texture$\;\to\;$Shade&16.53 / 25.07 & 18.26 / 28.60& \textbf{15.53} / \textbf{23.77}&16.10 / 24.36 \\
Shade+Texture$\;\to\;$RGB
& 62.63 / 1.65$\;\;$& 73.27 / 2.33$\;\;$& \textbf{61.03} / \textbf{1.50}$\;\;$&{61.25} / {1.60}$\;\;$\\
\midrule

RGB+Depth$\;\to\;$SemSeg&21.52 / 36.24&22.80 / 37.09&\textbf{18.57} / \textbf{33.29}&18.71 / 33.56 \\
RGB+SemSeg$\;\to\;$Depth&4.63 / 8.59&5.03 / 8.81 & \textbf{4.02} / \textbf{7.90}&4.27 / 8.26\\
Depth+SemSeg$\;\to\;$RGB &99.60 / 4.18$\;\;$ &102.97 / 4.31$\;\;\;\,$ &\textbf{96.13} / \textbf{3.66}$\;\,$ &97.01 / 3.94$\;\,$\\
\midrule
RGB+Depth$\;\to\;$Normal& 13.03 / 28.75 & 15.72 / 31.15& 12.26 / 27.12&\textbf{11.94} / \textbf{26.79} \\
RGB+Normal$\;\to\;$Depth& 3.34 / 5.22 &  4.67 / 6.73& 2.63 / 4.70&\textbf{2.57} / \textbf{4.45}\\
Depth+Normal$\;\to\;$RGB & 84.33 / 2.70$\;\;$ &90.49 / 3.73$\;\;$ & \textbf{82.81} / \textbf{2.64}$\;\;$&83.73 / 2.66$\;\;$ \\
\midrule
RGB+Depth$\;\to\;$Curve&$\;\,$5.42 / 15.09&$\;\,$5.73 / 16.08&$\;\,$\textbf{4.83} / \textbf{13.71}& $\;\,$5.03 / 14.15\\
RGB+Curve$\;\to\;$Depth&2.62 / 3.87&2.82 / 4.23 & \textbf{2.14} / \textbf{3.47}&2.25 / 3.67\\
Depth+Curve$\;\to\;$RGB &85.13 / 2.82$\;\,$ &88.69 / 3.39$\;\,$&\textbf{83.85} / \textbf{2.42}$\;\,$&84.52 / 2.64$\;\,$ \\
\midrule
Depth+Normal$\;\to\;$Shade&$\;\,$7.10 / 11.22 & $\;\,$7.47 / 11.45  & $\;\,$7.03 / 10.65 & $\;\,$\textbf{6.60} / \textbf{10.31} \\
Shade+Depth$\;\to\;$Normal&13.11 / 31.57 &13.74 / 32.20& 13.12 / 31.65 & \textbf{12.92} / \textbf{31.30} \\
Shade+Normal$\;\to\;$Depth&1.62 / 2.91 &1.92 / 3.18& 1.56  / 2.94 & \textbf{1.50} / \textbf{2.87} \\
\midrule
Total params. (M)& Gen: 163.3; Dis: 8.3&Gen: 124.2; Dis: 8.3&Gen: 124.2; Dis: 8.3&Gen: \textbf{54.5}; Dis: 8.3\\
\bottomrule[1pt]
\end{tabular}}
\label{tabs:cycle}
\vskip -0.03 in
\end{table*}

\begin{figure*}[h]
\centering
\includegraphics[scale=0.135]{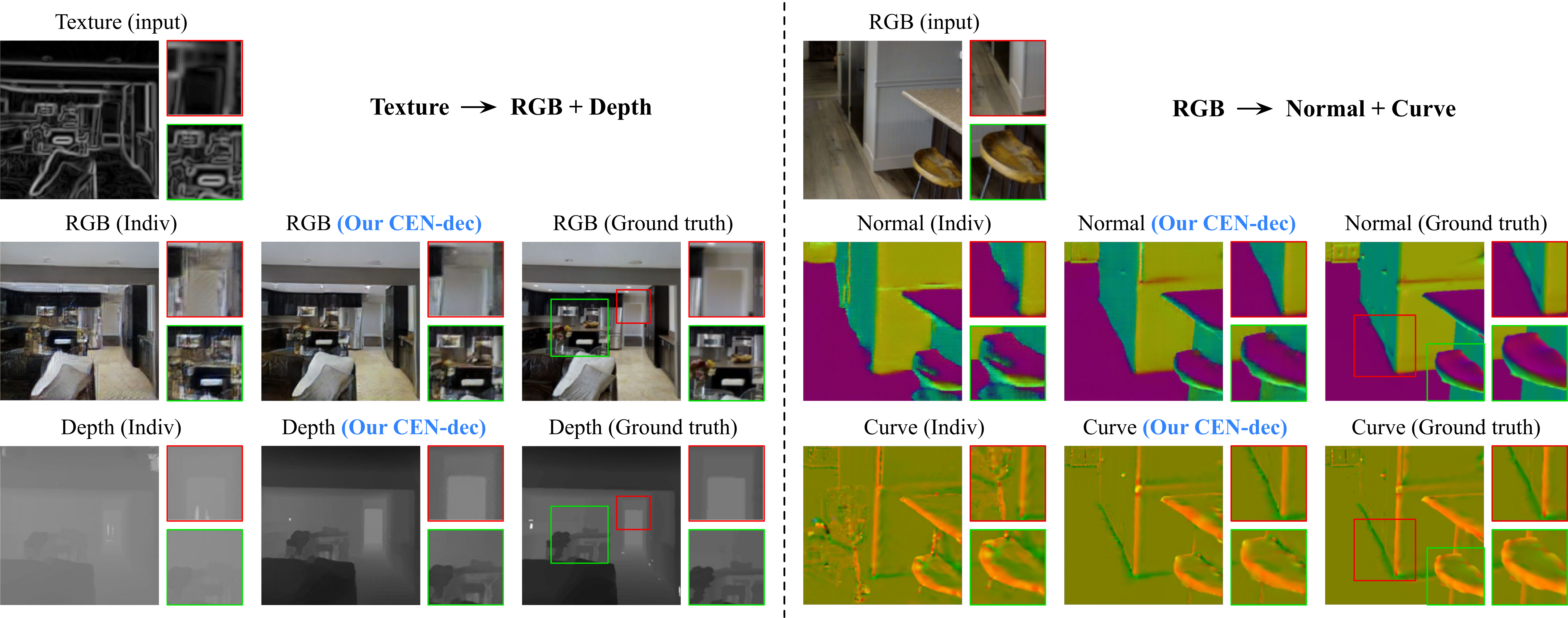}
\vskip-0.03in
\caption{Visualization results of multitask image-to-image translation including Texture$\to$RGB+Depth (left group) and RGB$\to$ Normal+Curve (right group), respectively. ``Curve'' is the abbreviation for the principle curve modality. We compare the individual (Indiv) baseline with unshared encoders  and our CEN-dec.
The resolution of each predicted image is $256\times256$. }
\vskip0.03in
\label{pic:trans_multitask}
\end{figure*}

\begin{table*}[t]
\centering
\caption{Experimental results of  multitask learning. Evaluation metrics are FID/KID ($\times 10^{-2}$) for RGB predictions and MAE ($\times 10^{-2}$)/MSE ($\times 10^{-2}$) for other predictions. Lower values indicate better performance. Individual (Indiv) learning and Cross-Task Consistency (X-TC)~\cite{DBLP:conf/cvpr/ZamirSCSCMG20} are served as baselines. We provide  numbers of groups for instance normalization (IN), encoder (enc) and decoder (dec), and the  total parameters (params.) in  generator (Gen) and discriminator (Dis), respectively. ``Curve'' and ``SemSeg'' are abbreviations for the principle curve and semantic segmentation, respectively.}
\tablestyle{1pt}{1.1}
\resizebox{150mm}{!}{
\begin{tabular}{c|p{2.4cm}<{\centering}p{2.4cm}<{\centering}p{2.4cm}<{\centering}p{2.4cm}<{\centering}p{2.8cm}<{\centering}c}
\toprule[1pt]
$\;$Modality&\makecell[c]{Indiv (IN$\times$2,\\enc$\times$2, dec$\times$2)}&\makecell[c]{Indiv (IN$\times$2,\\enc$\times$1, dec$\times$2)}&\makecell[c]{X-TC~\cite{DBLP:conf/cvpr/ZamirSCSCMG20} (IN$\times$2,\\enc$\times$1, dec$\times$2)} &\makecell[c]{CEN-dec (IN$\times$2,\\enc$\times$1, dec$\times$2)} &\makecell[c]{CEN-dec + X-TC~\cite{DBLP:conf/cvpr/ZamirSCSCMG20}\\(IN$\times$2, enc$\times$1, dec$\times$2)}\\
\midrule

\multirow{2}{*}{$ \text{RGB}\to\hskip-0.1em\left\{
\begin{aligned}
&\text{SemSeg} \\
&\text{Depth}\\
\end{aligned}
\right.$}&26.71 / 40.15&27.14 / 41.90&23.83 / 38.10&23.02 / 37.54 & \textbf{21.78} / \textbf{37.32} \\
&5.35 / 9.13&5.51 / 9.42&5.22 / 8.98& 4.82 / 8.50 & \textbf{4.76} / \textbf{8.43} \\
\midrule
\multirow{2}{*}{$ \text{RGB}\to\hskip-0.1em\left\{
\begin{aligned}
&\text{Normal} \\
&\text{Curve} \\
\end{aligned}
\right.$}
&18.74 / 37.24 &18.15 / 36.82&18.18 / 36.49& 16.74 / 32.26& \textbf{14.85} / \textbf{29.33}  \\$\;\,$
& $\;\,$6.24 / 16.97 & $\;\,$6.02 / 16.70 &$\;\,$5.33 / 14.76& $\;\,$4.92 / 14.08& $\;\,$\textbf{4.50} / \textbf{13.81}\\
\midrule
\hskip-0.16em\multirow{2}{*}{$ \text{RGB}\to\hskip-0.1em\left\{
\begin{aligned}
&\text{Shade}\\
&\text{Texture} \\
\end{aligned}
\right.$}& 24.04 / 33.85 &23.63 / 32.92 &19.04 / 29.87 & 18.77 / 27.94&\textbf{17.07} / \textbf{27.10} \\
&2.40 / 4.93 & 2.19 / 4.66 & 2.33 / 4.85 & 1.83 / 3.67& \textbf{1.64} / \textbf{2.99} \\

\midrule
\hskip-0.03em\multirow{2}{*}{$ \text{Texture}\to\hskip-0.1em\left\{
\begin{aligned}
&\text{RGB}\\
&\text{Depth} \\
\end{aligned}
\right.$}& 97.51 / 4.82$\;\,$& 96.81 / 4.57$\;\,$ &95.81 / 3.94$\;\,$& 92.92 / 3.25$\;\,$&\textbf{90.85} / \textbf{2.81}$\;\,$\\
&4.20 / 8.16 &4.05 / 7.94 & 3.54 / 6.07 & 3.19 / 5.05& \textbf{2.90} / \textbf{4.87} \\
\midrule
\hskip-0.03em\multirow{2}{*}{$ \text{Normal}\to\hskip-0.1em\left\{
\begin{aligned}
&\text{Depth}\\
&\text{Shade} \\
\end{aligned}
\right.$}& 2.59 / 3.92&2.72 / 4.16&2.20 / 3.54 & 1.97 / 3.30& \textbf{1.85} / \textbf{3.04}  \\
&$\;\,$8.08 / 12.40&$\;\,$7.90 / 12.03 &$\;\,$7.26 / 11.52 &$\;\,$7.09 / 11.14&$\;\,$\textbf{6.94} / \textbf{10.88} &\\

\midrule
Total params. (M)& Gen: 108.7; Dis: 8.3& Gen: 89.3; Dis: 8.3&Gen: 89.3; Dis: 8.3&Gen: 89.3; Dis: 8.3&Gen: 89.3; Dis: 8.3\\
\bottomrule[1pt]
\end{tabular}}
\vskip-0.05in
\label{tabs:multitask_cen}
\end{table*}

\begin{table*}[t]
\centering
\caption{Experimental results of simultaneously predicting three tasks. Evaluation metrics and abbreviations follow Table~\ref{tabs:multitask_cen}.  AdaShare (AS)~\cite{DBLP:conf/nips/SunPFS20} and Taskgrouping (TG)~\cite{DBLP:conf/icml/StandleyZCGMS20} are additionally served as  baselines.}
\tablestyle{1pt}{1.2}
\resizebox{150mm}{!}{
\begin{tabular}{c|p{2.4cm}<{\centering}p{2.4cm}<{\centering}p{2.4cm}<{\centering}p{2.4cm}<{\centering}p{3.2cm}<{\centering}c}
\toprule[1pt]
$\;$Modality&\makecell[c]{Indiv (IN$\times$3,\\enc$\times$3, dec$\times$3)}&\makecell[c]{AS~\cite{DBLP:conf/nips/SunPFS20} (IN$\times$1,\\enc$\times$2, dec$\times$3)}&\makecell[c]{TG~\cite{DBLP:conf/icml/StandleyZCGMS20} (IN$\times$2$\sim$3,\\enc$\times$2$\sim$3, dec$\times$3)}&\makecell[c]{CEN-dec (IN$\times$3,\\enc$\times$1, dec$\times$3)} & \makecell[c]{CEN-dec + TG~\cite{DBLP:conf/icml/StandleyZCGMS20}\\(IN$\times$3, enc$\times$2$\sim$3, dec$\times$3)}\\
\midrule

\hskip-0.6em\multirow{3}{*}{$ \text{RGB}\to\hskip-0.1em\left\{
\begin{aligned}
&\text{SemSeg} \\
&\text{Depth}\\
&\text{Normal}\\
\end{aligned}
\right.$}&26.68 / 40.11&29.50 / 43.71&26.72 / 40.15&25.30 / 39.64 & \textbf{22.97} / \textbf{37.50} \\
&5.35 / 9.15&5.02 / 8.71&5.15 / 8.80& 4.81 / 8.51 & \textbf{4.71} / \textbf{8.40} \\
&18.70 / 37.18&17.57 / 33.80&17.92 / 34.39& 17.05 / 33.19 & \textbf{16.63} / \textbf{32.02} \\

\midrule
\hskip-0.03em\multirow{3}{*}{$ \text{Texture}\to\hskip-0.1em\left\{
\begin{aligned}
&\text{RGB}\\
&\text{Depth}\\
&\text{Edge} \\
\end{aligned}
\right.$}& 97.45 / 4.80$\;\,$& 99.23 / 5.11$\;\,$ &97.40 / 4.78$\;\,$& 94.04 / 3.76$\;\,$&\textbf{92.71} / \textbf{3.20}$\;\,$\\
&4.24 / 8.19 &4.16 / 8.05 & 4.27 / 8.25 & 3.19 / 5.05& \textbf{3.08} / \textbf{4.96} \\
&0.97 / 1.73 &1.16 / 2.24 & 0.95 / 1.70 & 0.90 / 1.66& \textbf{0.86} / \textbf{1.61} \\
\midrule

\hskip-0.16em\multirow{3}{*}{$ \text{Normal}\to\hskip-0.1em\left\{
\begin{aligned}
&\text{RGB}\\
&\text{Depth}\\
&\text{Shade} \\
\end{aligned}
\right.$}
& 108.28 / 5.42$\;\;\;\,$&114.74 / 5.89$\;\;\;\,$&108.13 / 5.40$\;\;\;\,$& 102.55 / 5.20$\;\;\;\,$& \textbf{99.18} / \textbf{4.86} $\;$ \\
& 2.60 / 3.92&2.77 / 4.30&2.41 / 3.80 & 1.93 / 3.25& \textbf{1.83} / \textbf{3.01}  \\
&$\;\,$8.11 / 12.38&$\;\,$7.86 / 11.95 &$\;\,$7.75 / 11.77 &$\;\,$6.86 / 10.82&$\;\,$\textbf{6.79} / \textbf{10.73} &\\

\midrule
Total params. (M)& \makecell[c]{Gen: 163.1;\\Dis: 12.5}& \makecell[c]{Gen: 143.7;\\Dis: 12.5}& \makecell[c]{Gen: 143.7$\sim$163.1;\\Dis: 12.5}&\makecell[c]{Gen: 124.3;\\Dis: 12.5}& \makecell[c]{Gen: 143.7$\sim$163.1;\\Dis: 12.5}\\
\bottomrule[1pt]
\end{tabular}}
\label{tabs:multitask_cen2}
\end{table*}

\begin{table*}[t]
\centering
\caption{Experimental results of multimodal multitask learning. Evaluation metrics are FID/KID ($\times 10^{-2}$) for RGB predictions and MAE ($\times 10^{-2}$)/MSE ($\times 10^{-2}$) for other predictions. Lower values indicate better performance. Individual (Indiv) learning is served as the baseline. We provide  numbers of groups for instance normalization (IN), encoder (enc) and decoder (dec), and the  total parameters (params.) in  generator (Gen) and discriminator (Dis). ``Curve'' and ``SemSeg'' are abbreviations for the principle curve and semantic segmentation, respectively.}
\vskip-0.2em
\tablestyle{5pt}{1.1}
\resizebox{150mm}{!}{
\begin{tabular}{p{2.8cm}<{\centering}|p{2.7cm}<{\centering}p{2.4cm}<{\centering}p{2.4cm}<{\centering}p{2.8cm}<{\centering}p{2.8cm}<{\centering}p{2.8cm}<{\centering}}
\toprule[1pt]
$\;$Modality&\makecell[c]{Indiv (IN$\times$4,\\enc$\times$2, dec$\times$2)}&\makecell[c]{CEN-enc (IN$\times$4,\\enc$\times$1, dec$\times$2)}&\makecell[c]{CEN-dec (IN$\times$4,\\enc$\times$1, dec$\times$2)} &\makecell[c]{CEN-enc \& dec\\(IN$\times$4, enc$\times$1, dec$\times$2)} \\
\midrule

\multirow{2}{*}{$\;\; \left.
\begin{aligned}
&\text{RGB} \\
&\text{Depth}\\
\end{aligned}
\right\}
\to\hskip-0.1em\left\{
\begin{aligned}
&\text{SemSeg} \\
&\text{Curve}\\
\end{aligned}
\right.$}&26.86 / 40.24&21.17 / 36.05&25.22 / 39.36&\textbf{20.25} / \textbf{35.17} \\
&$\;\,$5.97 / 16.51&$\;\,$5.49 / 15.30&$\;\,$5.76 / 16.04&$\;\,$\textbf{5.27} / \textbf{14.93}\\

\midrule
\multirow{2}{*}{$\;\, \left.
\begin{aligned}
&\text{RGB} \\
&\text{Depth}\\
\end{aligned}
\right\}
\to\hskip-0.1em\left\{
\begin{aligned}
&\text{Nomal} \\
&\text{Shade}\\
\end{aligned}
\right.$}&18.68 / 37.11&13.54 / 29.03&16.81 / 32.75&\textbf{12.23} / \textbf{27.39} \\
&$\;\,$8.62 / 12.76&$\;\,$7.37 / 11.09&$\;\,$8.20 / 12.14&$\;\,$\textbf{7.08} / \textbf{10.91}\\

\midrule
\multirow{2}{*}{$\;\;\;\, \left.
\begin{aligned}
&\text{RGB} \\
&\text{Edge}\\
\end{aligned}
\right\}
\to\hskip-0.1em\left\{
\begin{aligned}
&\text{Depth} \\
&\text{Normal}\\
\end{aligned}
\right.$}&4.49 / 9.80&2.81 / 6.77&4.02 / 8.53&\textbf{2.47} / \textbf{6.33} \\
&16.56 / 33.40&13.28 / 29.32&15.14 / 32.72& \textbf{12.62} / \textbf{28.71}\\

\midrule
\multirow{2}{*}{$ \left.
\begin{aligned}
&\text{Texture} \\
&\text{Shade}\\
\end{aligned}
\right\}
\to\hskip-0.1em\left\{
\begin{aligned}
&\text{RGB} \\
&\text{Depth}\\
\end{aligned}\;\;\;
\right.$}&97.31 / 4.76$\;\,$&62.47 / 1.63$\;\,$&87.50 / 3.72$\;\,$&\textbf{60.26} / \textbf{1.57}$\;\,$ \\
&2.66 / 4.20&1.64 / 3.03&2.18 / 3.77& \textbf{1.58} / \textbf{2.94}\\

\midrule
Total params. (M)& Gen: 108.7; Dis: 8.3& Gen: 89.3; Dis: 8.3&Gen: 89.3; Dis: 8.3&Gen: 89.3; Dis: 8.3\\
\bottomrule[1pt]
\end{tabular}}
\label{tabs:multimodal_multitask}
\vskip -0.05 in
\end{table*}

\subsubsection{Image-to-Image Translation}

\textbf{Comparison with baseline fusion methods.}
We then evaluate the performance given five specific translation cases, including Shade+Texture$\to$RGB, Depth+Normal$\to$RGB, RGB+Shade$\to$Normal, RGB+Normal$\to$Shade and RGB+ Edge $\to$Depth. In addition to the three baselines used in semantic segmentation (Concat, Self-attention, Align),  we conduct an extra aggregation-based method by using the average operation. All baselines perform fusion under four different kinds of strategies: early (at the 1st Conv-layer), middle (the 4th Conv-layer), late (the 8th Conv-layer), and all-layer fusion. Our method yields much lower FID/KID or MAE/MSE than others, especially when predicting the  RGB modality, as  detailed in Table~\ref{tabs:translation}. These results support the benefit of our proposed idea once again.  

Main visualizations are provided in Fig. \ref{pic:trans_1}.
We observe that when predicting RGB given texture and shade, the prediction solely predicted from the texture is vague at boundary lines, while the prediction solely from the shade misses some opponents, \emph{e.g.} the pendant lamp, and is weak in predicting handrails. When fusing both input modalities, the concatenation method is uncertain in the regions where both modalities have disagreements. Alignment and self-attention are still weak in combining both modalities at details. Our results are clear at boundaries and fine-grained details. When predicting depth given RGB and edge, it is straightforward to find the benefits of multimodal fusion in this figure. The depth predicted by RGB is good at predicting numerical values, but is weak in capturing boundaries, which results in vague and curving boundaries. Oppositely, the depth predicted by the edge well captures boundaries, but is relatively weak in determining numerical values. The alignment fusion method is still weak in capturing boundaries. Both concatenation and self-attention methods are able to combine the advantages of both modalities, but numerical values are still obviously lower than the ground truth. All illustrations verify that our CEN achieves better performance compared to baseline methods. More visualizations and baseline settings are provided in the appendix.

\textbf{Considering more input modalities.}
We test whether our method is applicable to the case with more than two modalities. For this purpose, Table~\ref{tabs:multimodal} presents the results of image translation to RGB by inputting from one to four modalities of Depth, Normal, Texture, and Shade. It is observed that increasing the number of modalities improves the performance consistently, suggesting much potential of applying our method towards various cases.

\subsection{Evaluations on cycle multimodal fusion}
In this subsection, we evaluate CEN-cycle, a cycle multimodal fusion mode of CEN to simultaneously tackle three generation flows with a compact structure. As described in \textsection~\ref{sec:cycle_fusion}, in cycle multimodal fusion, we go through all 6 flows where each flow contains two input modalities and one output modality. The subnetwork is trained with all the three flows at each step. For each flow, our default setting is employing the encoders and decoders with shared convolution parameters but unshared INs. 
To demonstrate the benefit of CEN-cycle, we also implement these baselines in Table~\ref{tabs:cycle}: independent CEN that trains each flow separately, CEN-random that randomly samples one of the three flows per training step, and CEN-cycle with unshared decoders. 

We observe that compared with independent CEN, CEN-cycle with unshared decoders not only compacts the overall model but also achieves provably better prediction performance. By further sharing the decoders, CEN-cycle further reduces the model size (needing about $1/3$ parameter) and still yields better results than independent CENs. In addition, CEN-random is inferior to  independent CEN, probably because it is ineffective to balance the training between different flows if only one flow is trainable per step.  
In summary, the results here support that performing CEN-cycle is valuable, and it is able to reuse the information in different generation flows that involve overlapping input/output modalities by parameter sharing and joint training.

\subsection{Evaluations on multitask learning}
This subsection evaluates multitask learning which adopts a single modality as input and simultaneously predicts two or three different modalities. As introduced in~\textsection~\ref{sec:multitask_decoder}, CEN is conducted on the decoder side, abbreviated as CEN-dec. 

Table~\ref{tabs:multitask_cen} reports the case of predicting two modalities. Besides individual training with shared/unshared encoders, we consider a stronger baseline named Cross-Task Consistency (X-TC)~\cite{DBLP:conf/cvpr/ZamirSCSCMG20} under the triangle loss setting. X-TC basically enforces an addition supervision to let one predicted modality generate the other one. As observed, CEN-dec outperforms individual learning and X-TC in all tasks, and its performance is further promoted if used along with X-TC, showing the compatibility between CEN-dec and X-TC. 

In Fig.~\ref{pic:trans_multitask}, we further provide visualizations of multitask learning. We observe that by simultaneously predicting RGB and depth from texture, our CEN-dec predicts noticeably better results. By simultaneously normal and the principle curve from RGB, predicted normal boundaries of the table and wall are more accurate with CEN-dec.

We also consider the case of predicting three modalities. To this end, we implement two recent popular methods including AdaShare (AS)~\cite{DBLP:conf/nips/SunPFS20} and Task-Grouping (TG)~\cite{DBLP:conf/icml/StandleyZCGMS20} which consider multitask learning by parameter sharing. Table~\ref{tabs:multitask_cen2} summarizes  the experimental results. We find that both AS and TG usually achieve better accuracy than individual learning on some tasks (for example RGB$\rightarrow$Depth) but at the sacrifice of other tasks (RGB$\rightarrow$SemSeg), probably owing to the negative transfer. Yet, our CEN-dec, which simply shares the encoders with individual INs and performs channel exchanging in the decoder, outperforms all methods by noticeable margins in all tasks, supporting the superiority of channel exchanging for message fusion between different tasks. Interestingly, when combined with TG, the performance of CEN-dec is boosted remarkably, implying the flexibility of integrating our method with other techniques.

\subsection{Evaluations on multimodal multitask learning}
We  evaluate our multimodal multitask CEN as a combination of multimodal fusion and multitask learning, as shown in Table~\ref{tabs:multimodal_multitask}. We compare four different settings including individual training (Indiv), CEN on the encoder (CEN-enc), CEN on the decoder (CEN-dec), and CEN on both the encoder and decoder (CEN-enc \& dec). All the settings maintain four individual INs that correspond to the four different input-output combinations, respectively. 

Results indicate that performing CEN either on the encoder or decoder is  beneficial compared with the individual training baseline. Generally speaking, CEN-enc obtains more benefits compared with CEN-dec. This is natural as each input modality contains  complementary information for predicting each output modality, hence CEN-enc is particularly advantageous. But different output modalities might not be necessarily related, and as a result,  CEN-dec gains smaller improvement. As expected, combining CEN-enc and CEN-dec can further improve each of them and delivers the best performance in all considered cases.

\subsection{Discussions}
\label{sec:discussion}

\begin{figure}[t]
\centering
\vskip-0.3em
\hskip-0.5em
\includegraphics[scale=0.322]{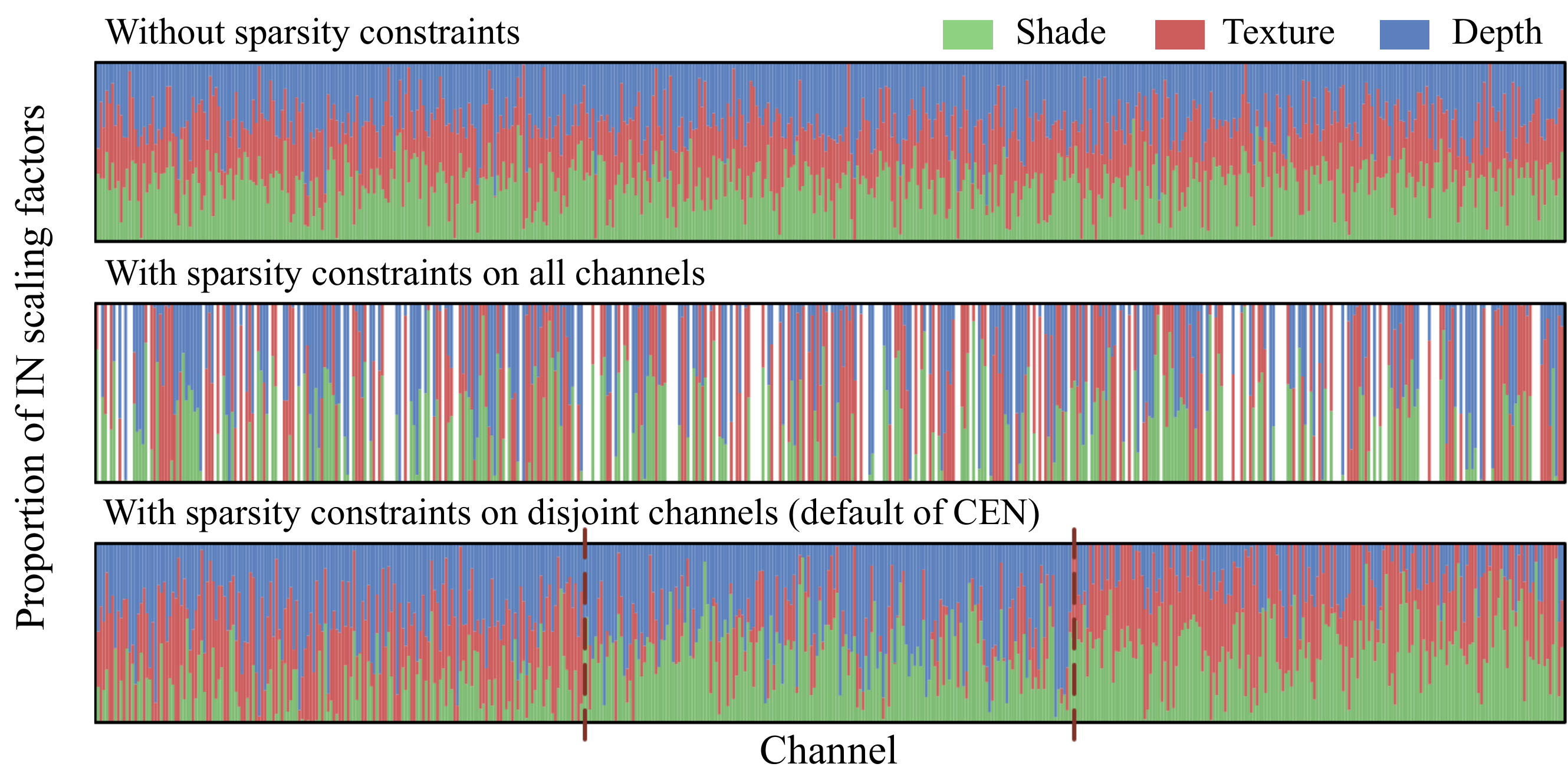}
\vskip-0.15em
\vskip-0.3em
\caption{We adopt shared Convs and unshared INs, and plot the proportion of scaling factors for each modality at the 7th Conv-layer, \emph{i.e.} $\gamma_{m,c}/(\gamma_{1,c}+\gamma_{2,c}+\gamma_{3,c})$, where $m=1,2,3$ being Shade, Texture and Depth, respectively. Note that we use the white space to represent a channel $c$ if  all of the three scaling factors ($\gamma_{1,c}$, $\gamma_{2,c}$, $\gamma_{3,c}$) are less than the threshold. }
\label{pic:featuremap_onelayer}
\vskip-0.8em
\vskip-0.8em
\end{figure}

\textbf{Why dividing channels into $M$ sub-parts.} We describe in \textsection~\ref{subsec:channel_exc} and Fig.~\ref{channel} that we evenly divide the whole channels into $M$ sub-parts (where $M$ is the number of input modalities), and apply  sparsity constraints only to one sub-part for each modality. Otherwise,  if we do not divide channels and apply sparsity constraints to all scaling factors for each modality, there is likely to be a portion of channels with close-to-zero scaling factors \emph{w.r.t.} all modalities. We provide the illustration in Fig.~\ref{pic:featuremap_onelayer}. We observe that with sparsity constraints on all channels, Fig.~\ref{pic:featuremap_onelayer} (middle) has a number of channels with small scaling factors, which are thus considered to be redundant \emph{w.r.t.} all modalities, which might lead to the decline of model capacity. Besides, it is hard to decide the exchanging direction on these redundant channels based on Eq.~\ref{eq:exchange-bn}. We provide corresponding experimental results in Table \ref{tabs:component} and the appendix (Table 13).

\begin{figure}[h!]
\centering
\hskip-0.05in
\includegraphics[scale=0.303]{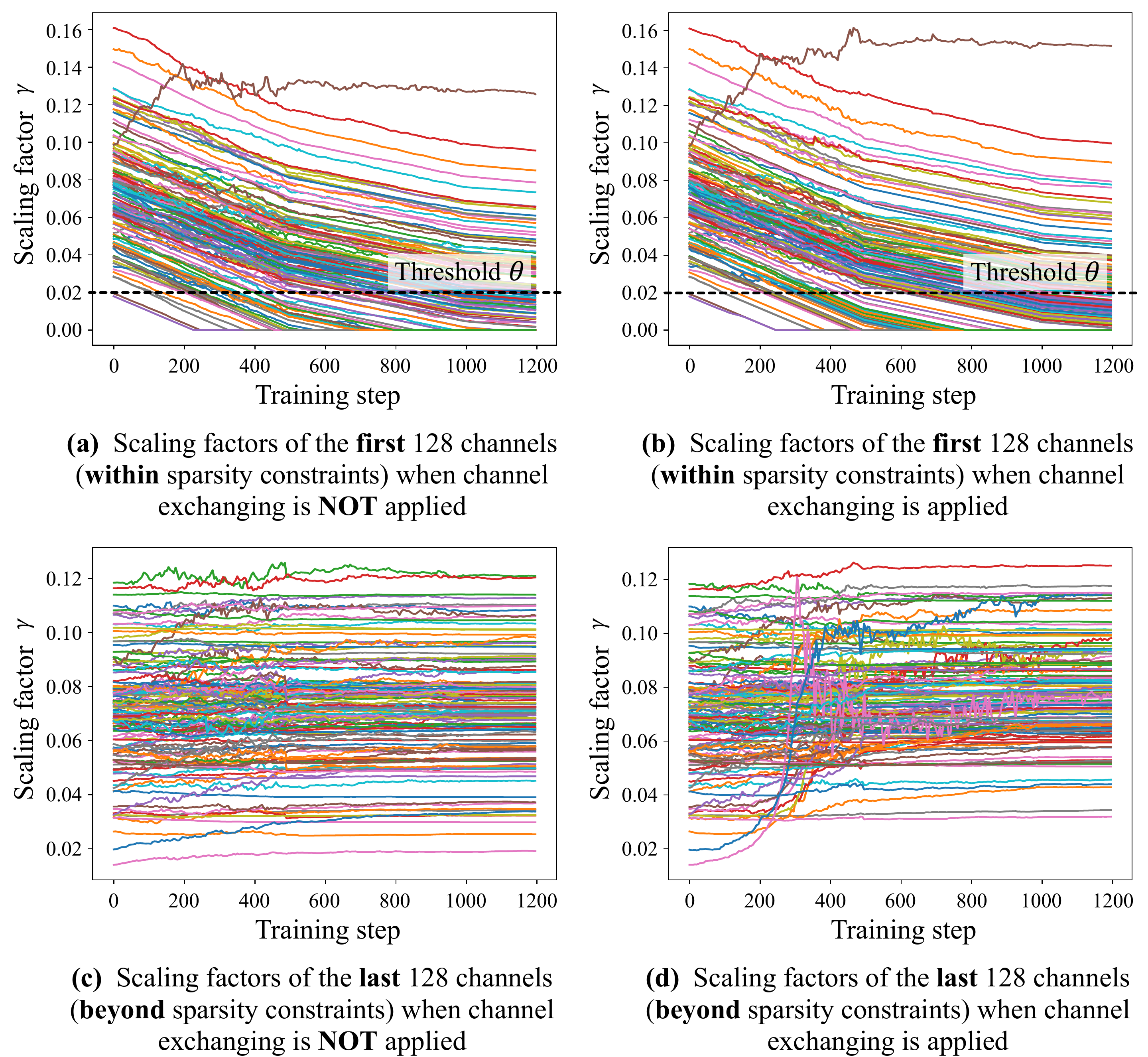}
\vskip0.02in
\caption{Typical values of  BN scaling factors (\emph{w.r.t.} the RGB modality) within/beyond sparsity constraints vs training steps. We  compare circumstances when channel exchanging is and is not applied. Experiments are conducted on NYUDv2 with RefineNet (ResNet101). We choose the 8th layer of convolutional layers that have $3\times3$ kernels, and there are  256 channels. Regarding  RGB,  sparsity constraints to  scaling factors are applied on the first 128 channels. }
\label{pic:scaling_factor}
\vskip-0.05in
\end{figure}

\textbf{Typical values of scaling factors.} Fig. \ref{pic:scaling_factor} demonstrates typical values of BN scaling factors vs training steps, consisting of four combinations: within/beyond sparsity constraints, and with/without channel exchanging. Experimental details are provided in the caption of Fig. \ref{pic:scaling_factor}. From the first two subfigures, we observe that whether applying channel exchanging or not,  scaling factors that are close to zero can hardly recover (in the later training process). In addition, according to the last two subfigures,  it seems channel exchanging increases the learning speed of a portion of scaling factors without sparsity constraints, probably due to the accumulated gradient on both the RGB branch and the depth branch by  channel exchanging (Eq.~\ref{eq:exchange-bn}).

\begin{figure}[t!]
\centering
\hskip-0.01in
\includegraphics[scale=0.163]{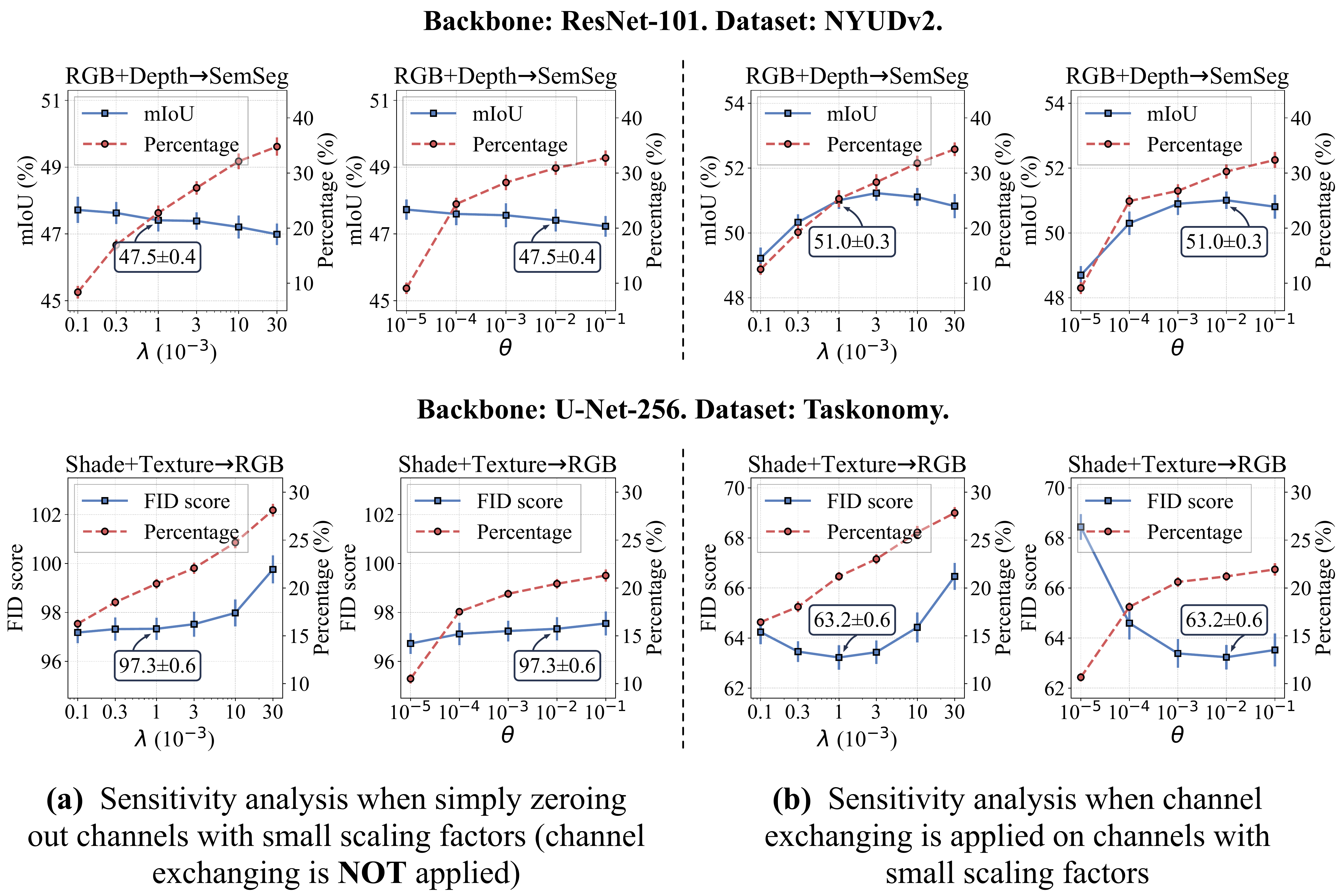}
\vskip0.02in
\caption{Effect when zeroing out channels (without channel exchanging) as well as the sensitivity analysis for $\lambda$ and $\theta$. Experiments include RGB-D semantic segmentation (SemSeg) on NYUDv2 (top group) and Texture+Shade$\to$RGB on Taskonomy (bottom group).  We conduct five experiments for each parameter setting. Default settings are  $\lambda=10^{-3}$ and $\theta=10^{-2}$. The left y-axis indicates the metric (mIoU $\uparrow$ or FID score $\downarrow$). The right y-axis indicates the percentage of channels that are lower than  $\theta$ and these channels will be replaced by zeros (left group) or by cross-modal channels (right group). Metric results at default settings are marked. }
\label{sec:params_sensitivity}
\vskip-0.1in
\end{figure}

\textbf{Effect of zeroing out channels and channel exchanging.} This part  provides the sensitivity analysis for two essential hyper-parameters of CEN, including the weight $\lambda$ (Eq.~\ref{eq:cen}) of sparsity constraint, and  the threshold $\theta$  (Eq.~\ref{eq:exchange-bn}) that identifies close-to-zero scaling factors. Experimental details are provided in the  caption of Fig.~\ref{sec:params_sensitivity}. To isolate the advantage of channel exchanging, Fig.~\ref{sec:params_sensitivity} (a) indicates that by zeroing out channels with small scaling factors (instead of channel exchanging), the performance slightly drops with the increase of $\lambda$ or $\theta$ since  the percentage of zeroing out channels increases accordingly. Nevertheless, such a drop is moderate, given that under the  sparsity constraints, the zeroed-out channels are less influential (as analyzed in \textsection~\ref{subsec:channel_exc}).  Fig.~\ref{sec:params_sensitivity} (b) provides the sensitivity analysis of our channel exchanging. We observe that both hyper-parameters $\lambda$ and $\theta$ are not sensitive around their default settings. It is also noticeable that without channel exchanging, simply zeroing out channels reaches much inferior performance.

\textbf{Importance of the exchanging process.} We  provide additional experiments in the appendix ({Table 12}), to evaluate the importance of the exchanging process. We try other approaches to  replace zeroed-out channels with: concatenated  multimodal features (followed by a Conv-layer) instead of the average,  evenly spaced channels from the same modality or other modalities, channels with the largest scaling factors, etc. Results indicate the superiority of our current design. In summary, albeit the simplicity of using the average of other modalities in CEN, it is also effective and competitive.

\textbf{Evaluation for the unsupervised learning.} In a part of multimodal fusion tasks, there is no ground truth during training~\cite{DBLP:journals/inffus/MaML19, DBLP:journals/pami/XuMJGL22,DBLP:journals/corr/abs-2205-07179}. As a general multimodal/multitask method,  CEN is also potentially applicable to unsupervised learning tasks. For example, we  apply CEN to the Saliency Network in \cite{DBLP:journals/corr/abs-2205-07179} for RGB-D unsupervised saliency detection,  a dense image prediction task aiming to effectively find and segment the most distinctive objects in a scene. Quantitive results and visualizations are provided in the appendix (Fig. 15 and Table 14), where  improvements are also achieved,  indicating the effectiveness of CEN in this case.

\section{Conclusion}
\vskip -0.03 in
We propose Channel-Exchanging-Network (CEN), a novel framework for multimodal fusion and multitask learning, which is parameter-free and self-adaptive. The motivation behind this is to boost inter-subnetwork fusion while simultaneously keeping sufficient intra-subnetwork processing. The channel exchanging is self-guided by channel importance measured by individual BNs, making our framework self-adaptive and compact. Extensive evaluations in four cases (multimodal fusion, cycle multimodal fusion, multitask learning, and multimodal multitask learning) verify the effectiveness of our method.

\appendices

\section{Implementation Details}
\label{sec:details}

In our experiments, we adopt ResNet101 and ResNet152 for semantic segmentation, and U-Net-256 for image-to-image translation. We use an NVIDIA Tesla V100 with 32GB for each experiment. Regarding both ResNet structures, we apply sparsity constraints on Batch-Normalization (BN) scaling factors \emph{w.r.t.} each Convolutional-layer (Conv-layer) with $3\times3$ kernels. These scaling factors further guide the channel exchanging process that exchanges a portion of feature maps after BN. For the Conv-layer with $7\times7$ kernels at the beginning of ResNet, and all other Conv-layers with $1\times1$ kernels, we do not apply sparsity constraints or channel exchanging. For U-Net, we apply sparsity constraints on Instance-Normalization (IN) scaling factors \emph{w.r.t.} all Conv-layers (eight layers in total) in the encoder of the generator, and each is followed by channel exchanging.

We mainly adopt three multimodal fusion baselines in our paper, including concatenation, alignment, and self-attention. Regarding the concatenation method, we stack multimodal feature maps along the channel, and then add a $1\times1$ convolutional layer to reduce the number of channels back to the original number. The alignment fusion method is a re-implementation of \cite{conf/eccv/WangWTSW16}, and we follow its default settings for hyper-parameter, \emph{e.g.} using 11 kernel functions for the multiple kernel Maximum Mean Discrepancy. The self-attention method is a re-implementation of the SSMA block proposed in \cite{journals/ijcv/ValadaMB20}, where we also follow the default settings, \emph{e.g.} setting the channel reduction ratio $\eta$ to 16.

In Table 2 of our main paper, we adopt early, middle, late and all-stage fusion for each baseline method. In ResNet101, there are four stages with 3, 4, 23, and 3 blocks, respectively. The early fusion, middle fusion, and late fusion refer to fusing after the 2nd stage, 3rd stage, and 4th stage respectively. All-stage fusion refers to fusing after the four stages.

\begin{table}[h!]
\centering
\caption{We compare training multimodal features in a parallel manner with different parameter sharing settings. Results of the proposed fusion method are reported in the last column. Evaluation metrics are FID/KID ($\times 10^{-2}$). We observe that the convolutional layers can be shared as long as we leave individual INs for different modalities, achieving even better performance.}
\tablestyle{2.5pt}{1.2}
\resizebox{0.98\linewidth}{!}{
\begin{tabular}{l|l|ccc|cccc}
\toprule[1pt]
Modality&\makecell[l]{Network\\stream}&\makecell[c]{Unshared Convs\\unshared INs}&\makecell[c]{Shared Convs\\shared INs}&\makecell[c]{Shared Convs\\unshared INs}&\makecell[c]{Multi-modal\\fusion}\\
\midrule
\multirow{3.5}{*}{\makecell[l]{\;\;Shade\\+Texture\\$\to$RGB}} 
& Shade&{102.21 / 5.25}&{112.40 / 5.58}&{100.69 / 4.51}&{72.07 / 2.32}\\
& Texture&{\;\;98.19 / 4.83}&{102.28 / 5.22}&{\;\;93.40 / 4.18}&{65.60 / 1.82}\\
\cmidrule(r){2-6}
&Ensemble&{\;\;92.72 / 4.15}&{\;\;96.31 / 4.36}&{\;\;87.91 / 3.73}&{62.63 / 1.65}\\
\midrule
\multirow{4.5}{*}{\makecell[l]{\;\;Shade\\+Texture\\+Depth\\$\to$RGB}} 
& Shade&{101.86 / 5.18}&{115.51 / 5.77}&{\;\;98.49 / 4.07}&{69.37 / 2.21}\\
& Texture&{\;\;98.60 / 4.89}&{104.39 / 4.54}&{\;\;95.87 / 4.27}&{64.70 / 1.73}\\
& Depth&{114.18 / 5.71}&{121.40 / 6.23}&{107.07 / 5.19}&{71.61 / 2.27}\\
\cmidrule(r){2-6}
&Ensemble&{\;\;91.30 / 3.92}&{100.41 / 4.73}&{\;\;84.39 / 3.45}&{58.35 / 1.42}\\
\midrule
\multirow{5.5}{*}{\makecell[l]{\;\;Shade\\+Texture\\+Depth\\+Normal\\$\to$RGB}} 
& Shade&{100.83 / 5.06}&{131.74 / 7.48}&{\;\;96.98 / 4.23}&{68.70 / 2.14}\\
& Texture&{\;\;97.34 / 4.77}&{109.45 / 4.86}&{\;\;94.64 / 4.22}&{63.26 / 1.69}\\
& Depth&{114.50 / 5.83}&{125.54 / 6.48}&{109.93 / 5.41}&{70.47 / 2.09}\\
& Normal&{108.65 / 5.45}&{113.15 / 5.72}&{\;\;99.38 / 4.45}&{67.73 / 1.98}\\
\cmidrule(r){2-6}
&Ensemble&{\;\;89.52 / 3.80}&{102.78 / 4.67}&{\;\;86.76 / 3.63}&{57.19 / 1.33}\\
\bottomrule[1pt]
\end{tabular}}
\label{tabs:supp_multimodal}
\end{table}

\begin{table}[t]
\centering
\caption{An Instance-Normalization layer consists of four components, including scaling factors $\bm{\gamma}$, offsets $\bm{\beta}$, running mean $\bm{\mu}$ and variance $\bm{\sigma}$. Following Table 5, we further compare the evaluation results with only unshared $\bm{\gamma},\bm{\beta}$, or only unshared $\bm{\mu},\bm{\sigma}$. Evaluation metrics are FID/KID ($\times 10^{-2}$). We observe that using unshared scaling factors and offsets seems to be more important. $\ell_1$ regulation and channel exchanging are not applied throughout these experiments.}
\resizebox{0.98\linewidth}{!}{
\tablestyle{2pt}{1.2}
\begin{tabular}{c|l|cc|cc}
\toprule[1pt]
Modality&\makecell[l]{Network\\stream}&\makecell[c]{Unshared Convs\\unshared INs}&\makecell[c]{Shared Convs\\unshared INs}&\makecell[c]{Shared Convs,$\bm{\gamma}$,$\bm{\beta}$\\unshared $\bm{\mu}$,$\bm{\sigma}$}&\makecell[c]{Shared Convs,$\bm{\mu}$,$\bm{\sigma}$\\unshared $\bm{\gamma}$,$\bm{\beta}$}\\
\midrule
\multirow{4.5}{*}{\makecell[l]{\;\;Shade\\+Texture\\+Depth\\$\to$RGB}} 
& Shade&{101.86 / 5.18}&{\;\;98.49 / 4.07}&{107.86 / 5.53}&{105.29 / 5.29}\\
& Texture&{\;\;98.60 / 4.89}&{\;\;95.87 / 4.27}&{105.46 / 5.25}&{102.90 / 5.06}\\
& Depth&{114.18 / 5.71}&{102.07 / 4.89}&{118.35 / 6.07}&{114.35 / 5.80}\\
\cmidrule(r){2-6}
&Ensemble&{\;\;91.30 / 3.92}&{\;\;84.39 / 3.45}&{\;\;96.30 / 4.41}&{\;\;92.25 / 4.02}\\
\midrule
\multirow{5.5}{*}{\makecell[l]{\;\;Shade\\+Texture\\+Depth\\+Normal\\$\to$RGB}} 
& Shade&{100.83 / 5.06}&{\;\;96.98 / 4.23}&{113.56 / 5.65}&{102.74 / 5.17}\\
& Texture&{\;\;97.34 / 4.77}&{\;\;94.64 / 4.22}&{105.36 / 5.32}&{\;\;97.53 / 4.56}\\
& Depth&{114.50 / 5.83}&{109.93 / 5.41}&{119.31 / 6.20}&{112.73 / 5.60}\\
& Normal&{108.65 / 5.45}&{\;\;99.38 / 4.45}&{108.01 / 5.06}&{100.34 / 4.53}\\
\cmidrule(r){2-6}
&Ensemble&{\;\;89.52 / 3.80}&{\;\;86.76 / 3.63}&{\;\;95.56 / 4.64}&{\;\;89.26 / 3.91}\\
\bottomrule[1pt]
\end{tabular}}
\label{tabs:supp_unshare_mn}
\vspace{-1em}
\end{table}

\begin{table*}[h]
	\centering
	\caption{Additional experiments on the NYUDv2 dataset based on  RefineNet (ResNet101) to evaluate the importance of the exchanging process. Results include multimodal fusion on image translation (to RGB) with   $2\sim4$ input modalities. Evaluation metrics are FID/KID ($\times 10^{-2}$) and lower values indicate better performance.
	} 
	\vskip -0.02in
	\label{table:replace}
	\tablestyle{1pt}{1.3}
	\resizebox{1\linewidth}{!}{
		\begin{tabular}{l|c|c|cccc}
		\toprule[1pt]
			\label{tab:possiblemethods}
 For modality $m$, replacing (the feature map) at a zeroed-out channel (channel index $i$) with:
			& Depth+Normal
			& \makecell[l]{Depth+Normal\\+Texture}&\makecell[l]{Depth+Normal\\+Texture+Shade}\\
			\midrule
			A zero embedding (only fusion by ensemble) & 107.32 / 5.39 & 96.90 / 4.75 & 95.50 / 4.68\\
			The  $i$-th channel from another one  modality $m'\ne m$&Same with CEN & 63.14 / 1.73 & 62.76 / 1.69\\
			 Average of  evenly spaced channels \textbf{[a]} (Fig.~\ref{pic:replace}) (beyond sparsity constraints) from the same modality $m$& 106.62 / 5.29 & 95.63 / 4.64 & 95.90 / 4.71 \\
			One random channel in the region \textbf{[b]} (Fig.~\ref{pic:replace}) (beyond sparsity constraints) from the same modality $m$& 109.71 / 5.62 & 97.06 / 4.92 & 96.64 / 4.85 \\
			Average of  evenly spaced channels \textbf{[c]} (Fig.~\ref{pic:replace}) from  other modalities $\forall m'\ne m$ & 89.52 / 3.56 & 68.11 / 2.06 & 66.15 / 1.91\\
			Average of channels including both \textbf{[c]} and \textbf{[d]} (Fig.~\ref{pic:replace})  from  other modalities $\forall m'\ne m$ &85.08 / 2.92 & 64.73 / 1.82 & 61.99 / 1.64 \\
			Average of unused channels with the  largest scaling factors from other modalities $\forall m'\ne m$ & 88.61 / 3.13& 68.09 / 2.10 & 68.87 / 2.15\\
			Weighted average of the $i$-th channels  \textbf{[d]} (Fig.~\ref{pic:replace}) from  other modalities $\forall m'\ne m$ &Same with CEN & 61.07 / 1.59 & 58.26 / 1.40 \\
			Concatenation (followed by a $1\times1$ Conv) of the $i$-th channels \textbf{[d]} (Fig.~\ref{pic:replace}) from  other modalities $\forall m'\ne m$ & Same with CEN & 63.32 / 1.73 & 61.70 / 1.64\\
			\rowcolor{gray!10}
			Average of the $i$-th channels \textbf{[d]} (Fig.~\ref{pic:replace}) from  other modalities $\forall m'\ne m$ (our CEN) & \textbf{84.33} / \textbf{2.70} & \textbf{60.90} / \textbf{1.56} & \textbf{57.19} / \textbf{1.33}\\
			\bottomrule[1pt]
		\end{tabular}
	}
\vskip -0.03in
\end{table*}

\begin{figure}[h!]
\centering
\vskip-0.05in
\includegraphics[scale=0.8]{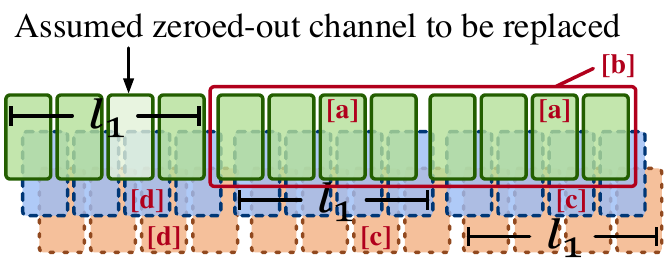}
\caption{Illustrations of  channels as a complement to Table~\ref{table:replace}.}
\label{pic:replace}
\end{figure}

We now introduce the metrics (including FID and KID) used in our image-to-image translation task.

Firstly,  Fréchet-Inception-Distance (FID) \cite{conf/nips/HeuselRUNH17} mainly contrasts the statistics of generated samples against real samples.  FID fits a Gaussian distribution to the hidden activations of InceptionNet for each compared image set and then computes the Fréchet distance (also known as the Wasserstein-2 distance) between those Gaussians. Lower FID is better, indicating that the generated images are more similar to the real ones.

Secondly, Kernel-Inception-Distance (KID) ~\cite{conf/iclr/BinkowskiSAG18} is a metric similar to the FID score but uses the squared Maximum-Mean-Discrepancy (MMD) between Inception representations with a polynomial kernel. Unlike FID, KID has a simple unbiased estimator, making it more reliable especially when there are much more inception feature channels than image numbers. Lower KID indicates more visual similarity between real and generated images. Regarding our implementation of KID, the hidden representations are derived from the Inception-v3 pool3 layer.

\begin{figure}[t!]
\centering
\hskip-0.03in
\includegraphics[scale=0.257]{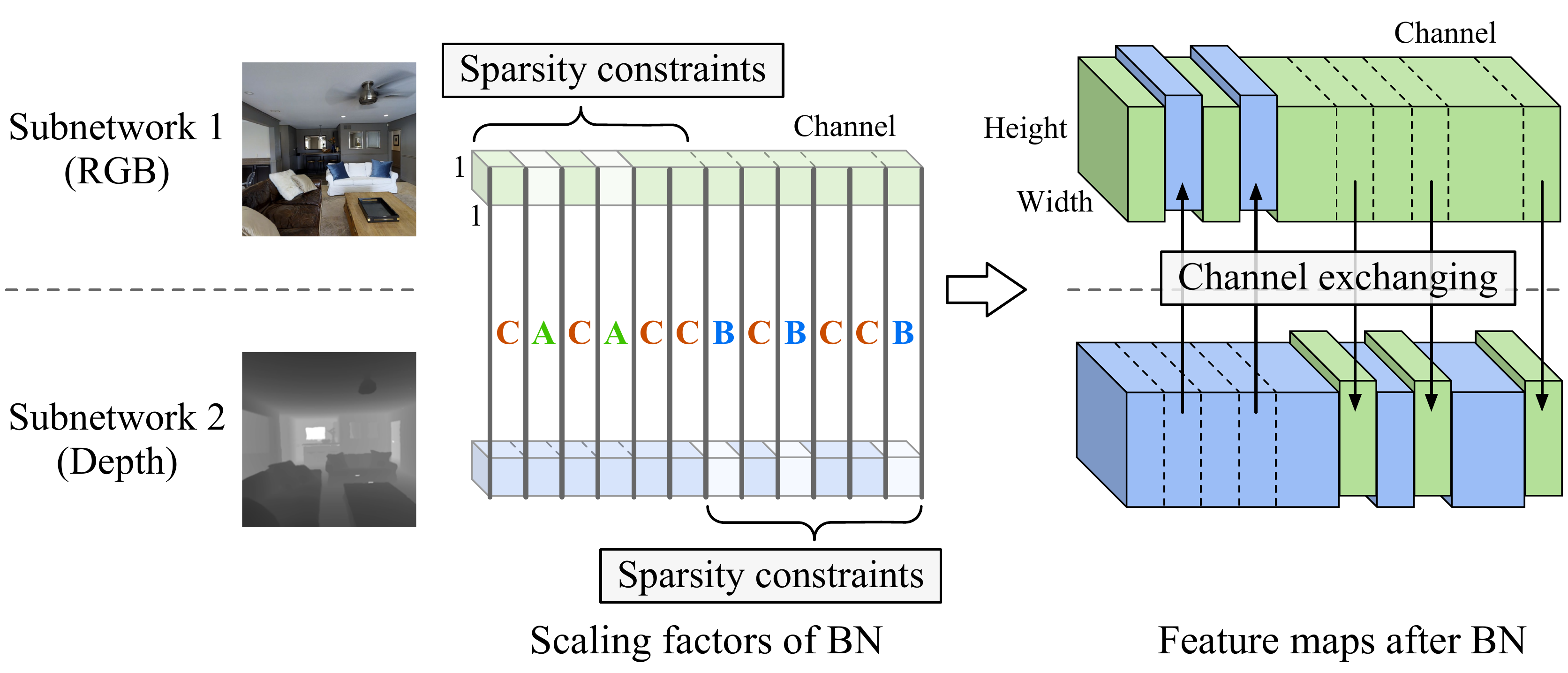}
\caption{An illustration of CEN. The sparsity constraints on scaling factors are applied to disjoint channel regions of different modalities. As annotated, each channel is categorized to  \textbf{A}, \textbf{B}, or \textbf{C}, based on its $\gamma_{rgb}$ and $\gamma_{depth}$.}
\label{channel}
\vskip -0.05 in
\end{figure}

\begin{figure}[t]
\centering
\vskip-0.05in
\hskip-0.5em
\includegraphics[scale=0.39]{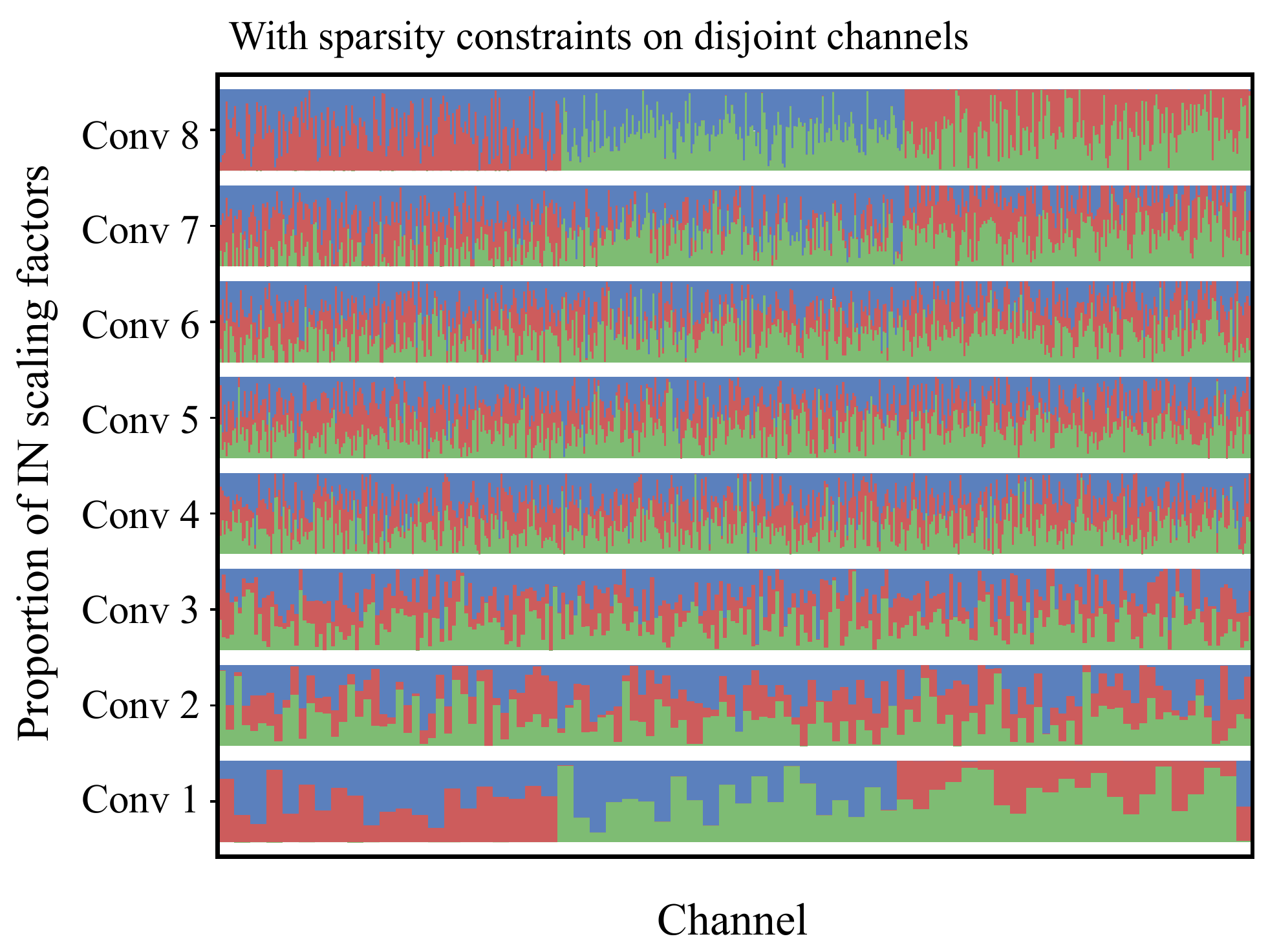}
\vskip-0.15em
\vskip-0.03in
\caption{We adopt shared Convs and unshared INs, and plot the proportion of scaling factors for each modality at each Conv-layer, \emph{i.e.} $\gamma_{m,c}/(\gamma_{1,c}+\gamma_{2,c}+\gamma_{3,c})$, where $m=1,2,3$ being Shade, Texture and Depth, respectively.Proportion of scaling factors in all Conv-layers, and sparsity constraints are applied on disjoint channels.} 
\label{pic:featuremap_onelayer}
\end{figure}

	\begin{table}[h!]
\centering
\caption{Multimodal fusion on image translation (to RGB) with  or without (w/o) dividing channels into $M$ sub-parts. Evaluation metrics are FID/KID ($\times 10^{-2}$). Lower values indicate better performance for both metrics.}
\vskip-0.02in
\tablestyle{2pt}{1.3}
\resizebox{86mm}{!}{
\begin{tabular}{c|cccc|ccc}
\toprule[1pt]
Method&\makecell[l]{Depth+Normal}&\makecell[l]{Depth+Normal\\+Texture}&\makecell[l]{Depth+Normal\\+Texture+Shade}\\
\midrule
Dividing $M$ sub-parts (default) &84.33 / 2.70&60.90 / 1.56&57.19 / 1.33\\
W/o dividing $M$ sub-parts & 87.63 / 3.49 & 65.12 / 1.90 & 64.87 / 1.85\\
\bottomrule[1pt]
\end{tabular}}
\label{tabs:multimodal-divide}
 \vskip -0.1 in
\end{table}

\begin{figure*}[h]
\centering
\includegraphics[scale=0.14]{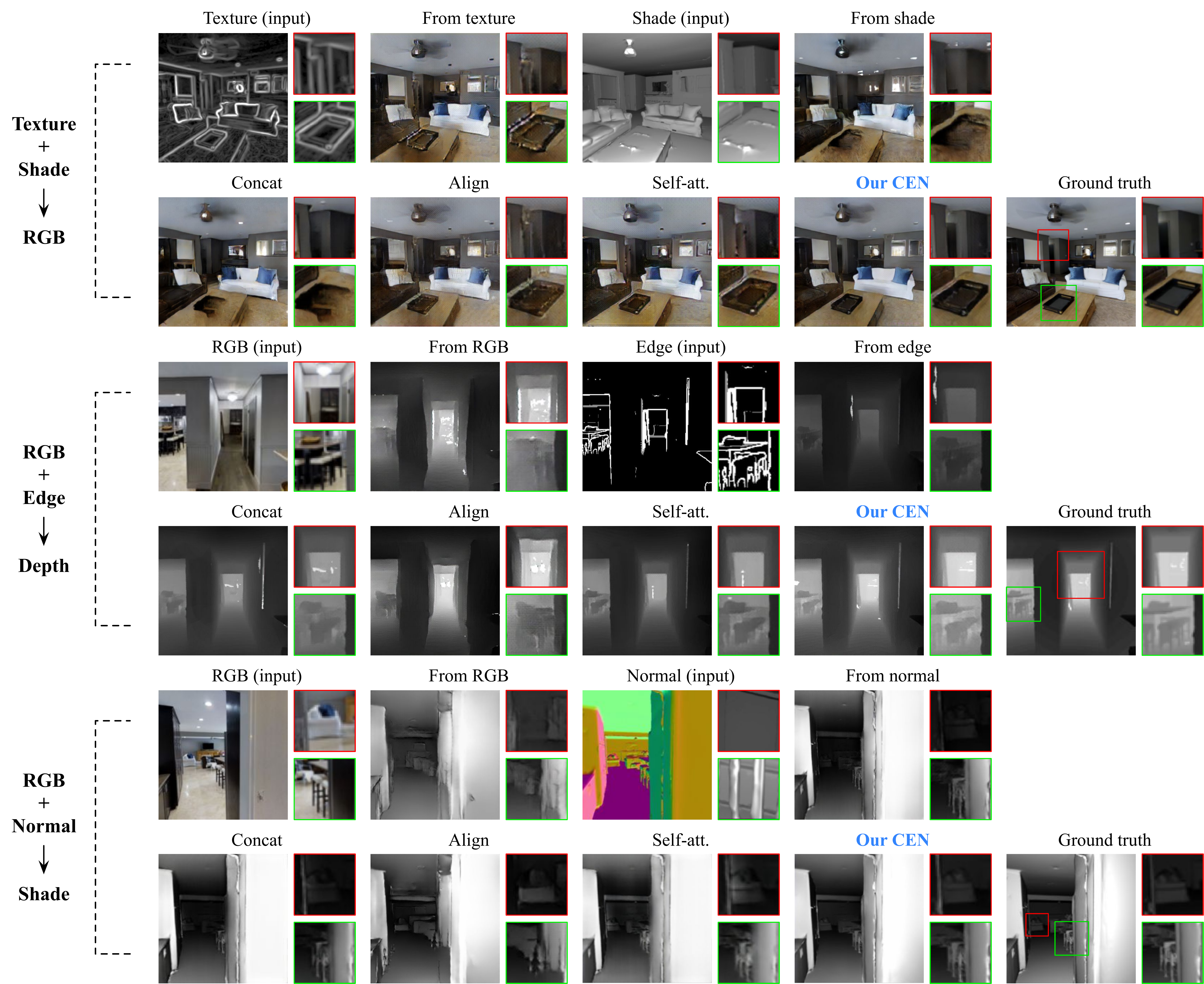}
\caption{Additional visualization results of image-to-image translation including Texture+Shade$\to$RGB (top group), RGB+Edge$\to$Depth (middle group), and RGB+Normal$\to$Shade (bottom group), respectively.  The resolution of each predicted image is $256\times256$. }
\label{pic:trans_2}
\vspace{-0.5em}
\end{figure*}

\begin{figure*}[h]
\centering
\vskip-0.05in
\includegraphics[width=0.72\linewidth]{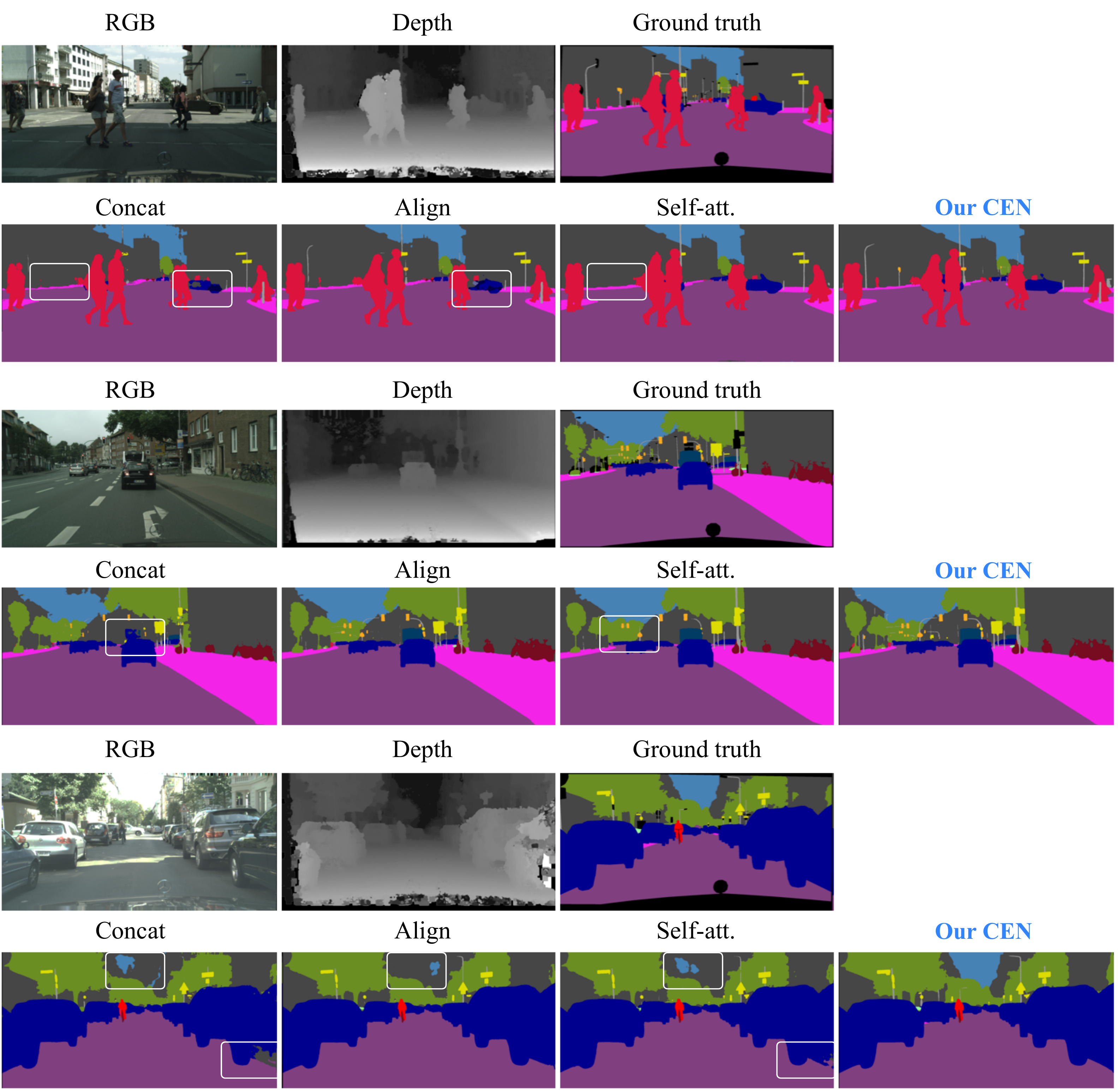}
\caption{Visualization  for the semantic segmentation on  Cityscapes  \cite{conf/cvpr/CordtsORREBFRS16}. For the baseline methods, we use white frames to highlight the regions with poor prediction results. We  observe that when the light intensity is high,  baseline methods are weak in capturing the boundary between the sky and buildings using the depth information. Besides, the concatenation and self-attention methods do not preserve fine-grained objects, \emph{e.g.} traffic signs, and are sensitive to noises of the depth input (see the rightmost vehicle in the last group). In contrast, the prediction of our CEN is better in these  aspects.}
\label{pic:city_seg_results}
\end{figure*}

\begin{figure}[t!]
\centering
\hskip-0.04in
\includegraphics[scale=0.31]{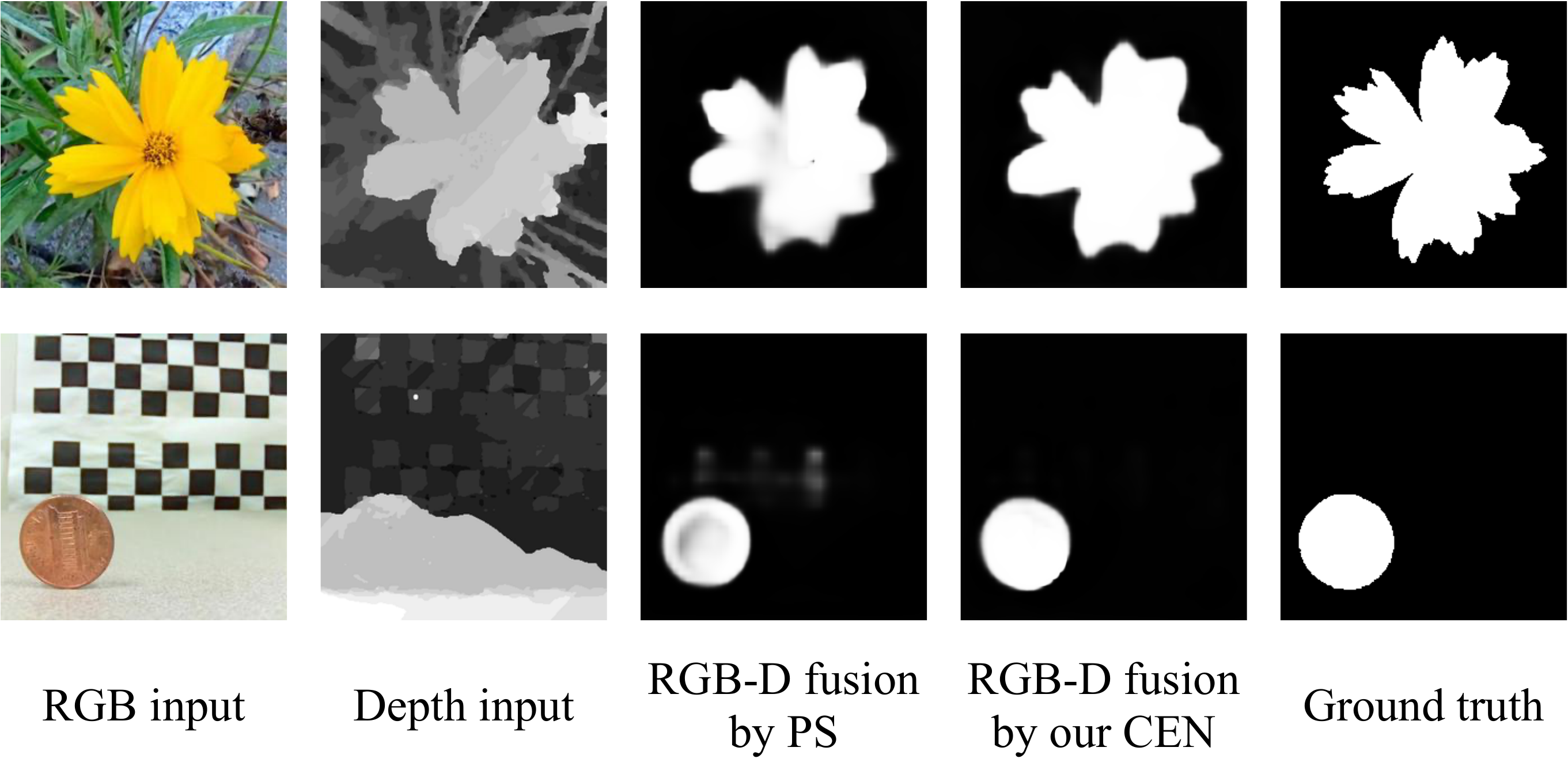}
\vskip -0.05 in
\caption{Visualization of applying CEN to the unsupervised RGB-D saliency detection. We compare our method with another RGB-D-based method Promoting Saliency (PS)~\cite{DBLP:journals/corr/abs-2205-07179} which recently achieves SOTA.}
\label{pic:silency}
\vskip -0.15 in
\end{figure}

\section{Additional Discussions and Results}

\textbf{Dividing channels into sub-parts.}
Here we provide additional descriptions of why dividing channels into $M$ sub-parts and individually applying sparsity constraints. We first use the case with 2 modalities for example. As shown in Fig.~\ref{channel}, we divide channels into 2 disjoint sub-parts and apply sparsity constraints. During training, each certain channel belongs to one of the three  categories: \textbf{A} (where $\gamma_{rgb}\approx 0, \gamma_{depth}> 0$), \textbf{B} (where $\gamma_{rgb}> 0, \gamma_{depth}\approx 0$), and \textbf{C} (where $\gamma_{rgb}> 0, \gamma_{depth}> 0$). There won't be $(\gamma_{rgb}\approx 0, \gamma_{depth}\approx 0)$ as we apply sparsity constraints on disjoint sub-parts. However, if we apply sparsity constraints on all scaling factors for each modality (without dividing 2 sub-parts), there is likely be a portion of channels with close-to-zero scaling factors \emph{w.r.t.} both modalities, \emph{i.e.}, $(\gamma_{rgb}\approx 0, \gamma_{depth}\approx 0)$. These channels are considered to be unimportant/redundant for both modalities. Regarding multimodal fusion, it is kind of waste of channels, which might lead to the decline of model capacity. Besides, it is hard to decide the exchanging direction on these channels according to Eq. 4 (main paper).

Similarly, when there are 3 (or $M$) modalities as input, dividing the whole channels into 3 (or $M$) sub-parts avoids a channel from being redundant for all modalities. As a result, we divide channels into $M$ sub-parts and apply sparsity constraints on one sub-part for each modality. 
 As an example, for Shade+Texture+Depth$\to$RGB image-to-image translation with shared Convs and unshared INs, channels are evenly divided into three sub-parts. We plot the proportion of IN scaling factors at  each Conv-layer in the encoder of U-Net in Fig. \ref{pic:featuremap_onelayer}.

Comparison results to support the channel dividing  have been provided in Table 1 of our main paper: semantic segmentation ``All-channel'' (49.8) vs ``Half-channel'' (51.1). Additional  results for image-to-image translation are shown in Table~\ref{tabs:multimodal-divide}. All these results indicate the superiority of applying sparsity constraints on sub-parts of channels.

\textbf{Effect of network sharing. } In Table \ref{tabs:supp_multimodal}, we verify that sharing convolutional layers (Convs) but using individual Instance-Normalization layers (INs) allows 2$\sim$4 modalities trained in a single network, and even achieve better performance than training with individual networks. Again, if we further share INs, there will be an obvious performance drop. More detailed comparison is provided in Table \ref{tabs:supp_unshare_mn}.

\textbf{Visualization of indoor experiments.} We provide additional visualizations of the image-to-image translation task in Fig. \ref{pic:trans_2}, as a complement to Fig. 5 (main paper). Regarding baseline implementation in all these visualizations, we adopt all-layer fusion (fusion at all eight Conv-layers in the encoder) for concatenation and self-attention methods, and adopt middle fusion (fusion at the 4th Conv-layer) for the alignment method. These settings achieve relatively high performance regarding baseline methods according to their numerical results. 

\textbf{Visualization of outdoor experiments.} In this part, we additionally conduct outdoor semantic segmentation experiments on the Cityscapes dataset \cite{conf/cvpr/CordtsORREBFRS16} and provide the visualization comparison. Cityscapes is an outdoor dataset containing images from 27 cities in Germany and neighboring countries. The dataset contains 2,975 training, 500 validation and 1,525 test images. There are 20,000 additional coarse annotations provided by the dataset, which are not used for training in our experiments.  All results are obtained with the backbone PSPNet (ResNet101) of single-scale evaluation for test. These visualizations are provided in Fig~\ref{pic:city_seg_results}.

\begin{table}[h!]
\centering
\caption{Quantitave results of applying CEN to the unsupervised RGB-D saliency detection. We follow the training settings in~\cite{DBLP:journals/corr/abs-2205-07179}. Evaluation datasets include NJUD~\cite{NJU2K}, NLPR~\cite{NLPR}, STERE~\cite{STERE}, and DUTLF-Depth~\cite{DMRA2019}. We adopt Mean Absolute Error (MAE)~\cite{borji2015salient} as the evaluation metric following ~\cite{DBLP:journals/corr/abs-2205-07179}. Lower values indicate better performance.}
\vspace{-0.02in}
\tablestyle{10pt}{1.02}
\resizebox{86mm}{!}{
    \begin{tabular}{c|c|c|c|ccc}
      \toprule[1pt]
    Method&NJUD&NLPR&STERE&DUTLF-Depth\\
    \midrule
          MST~\cite{MST2016}    & .281 & .199 & .269 & .279 \\
          BSCA~\cite{BSCA2015}  & .216 & .178  & .179  & .181 \\
          GP~\cite{GP2015}  & .204 & .144 & .182  & - \\          
          CDB~\cite{CDB2018}   & .200	& .108	& .166  & - \\
          SE~\cite{SE2016}   & .164 & .085  & .143  & .196 \\
          DCMC~\cite{DCMC}   & .167 & .196 & .148 & .243 \\
          MB~\cite{MB2017}   & .202& .089 & .178  & .156 \\
          CDCP~\cite{CDCP2017}   & .181 & .114 & .149 & .159 \\
    USD~\cite{USD2018}    & .163    & .119    & .146   & .157 \\
          DeepUSPS~\cite{DeepUSPS2019}   & .159    & .088  & .124    & .149 \\       
          {SP}~\cite{DBLP:journals/corr/abs-2205-07179}  (RGB-D)   & {.135}  & {.065} & {.099}  & {.107} \\
     \midrule
     { CEN} (RGB-D)  & \textbf{.132}  & \textbf{.059} & \textbf{.095}  & \textbf{.103}\\
     \bottomrule[1pt] 
    \end{tabular}}
  \label{comUnSOD}
  \vspace{-0.2in}
\end{table}

\begin{table*}[h]
\centering
\vspace{0.2em}
\caption{Comparison on multimodal fusion, cycle multimodal fusion, multitask learning, and multimodal multitask learning with {15,000} sampled training images (instead of 1,000 for other  experiments). Evaluation metrics are FID/KID ($\times 10^{-2}$) for RGB predictions and MAE ($\times 10^{-2}$)/MSE ($\times 10^{-2}$) for other predictions. Lower values indicate better performance for all metrics.  ``Curve'', ``SemSeg'', and ``X-TC'' are abbreviations for the principle curve,  semantic segmentation, and Cross-Task Consistency~\cite{DBLP:conf/cvpr/ZamirSCSCMG20} respectively. More abbreviation details  follow the captions of Table 4, 6, 7, 9 in our paper.}
\vspace{0.4em}
\tablestyle{5pt}{1.1}
\resizebox{160mm}{!}{
\begin{tabular}{c|ccccccc}
\toprule[1pt]
\textbf{Multimodal fusion}&Concat &Average&Align&Self-att.&Our CEN\\
\midrule
\multirow{1}{*}{\makecell[l]{Shade+Texture$\;\to\;$RGB}} 
& 75.39 / 2.77 & 75.46 / 2.82 & 86.20 / 3.92 & 68.65 / 2.23 & \textbf{56.94} / \textbf{1.45}\\
\midrule
\multirow{1}{*}{\makecell[l]{Depth+Normal$\;\to\;$RGB}} &
91.64 / 3.38 &93.81 / 3.50& 97.05 / 3.99 & 88.60 / 3.02 & \textbf{79.68} / \textbf{2.59} \\
\midrule
\midrule

\textbf{Cycle multimodal fusion}&\makecell[c]{CEN (IN$\times$6, \\enc$\times$3, dec$\times$3)}&\makecell[c]{CEN-random (IN$\times$6, \\enc$\times$1, dec$\times$3)} &\makecell[c]{CEN-cycle (IN$\times$6, \\enc$\times$1, dec$\times$3)}&\makecell[c]{CEN-cycle (IN$\times$6, \\enc$\times$1, dec$\times$1)}\\
\midrule
RGB+Shade$\;\to\;$Texture& 1.59 / 2.80 & 1.94 / 4.07& \textbf{1.36} / \textbf{2.21}& {1.45} / {2.34}\\
RGB+Texture$\;\to\;$Shade&13.77 / 22.78 & 16.93 / 24.02& \textbf{12.91} / \textbf{21.77}&13.13 / 22.09 \\
Shade+Texture$\;\to\;$RGB
& 56.94 / 1.45$\;\;$& 66.51 / 2.18 $\;$& \textbf{55.26} / \textbf{1.38}$\;\;$&{56.03} / {1.43}$\;\;$\\
\midrule
RGB+Depth$\;\to\;$SemSeg&18.49 / 33.20 &20.03 / 35.44&\textbf{15.90} / \textbf{30.41}&16.72 / 31.03 \\
RGB+SemSeg$\;\to\;$Depth&3.72 / 7.54 &4.29 / 8.02 & \textbf{3.50} / \textbf{7.07}&3.54 / 7.31\\
Depth+SemSeg$\;\to\;$RGB &93.95 / 3.57$\;\;$ &96.40 / 3.65$\;\;$ &\textbf{91.72} / \textbf{3.30}$\;\,$ &92.44 / 3.51$\;\,$\\
\midrule
Total params. (M)& Gen: 163.3; Dis: 8.3&Gen: 124.2; Dis: 8.3&Gen: 124.2; Dis: 8.3&Gen: \textbf{54.5}; Dis: 8.3\\
\midrule
\midrule

\textbf{Multitask learning}&\makecell[c]{Indiv (IN$\times$2,\\enc$\times$2, dec$\times$2)}&\makecell[c]{Indiv (IN$\times$2,\\enc$\times$1, dec$\times$2)}&\makecell[c]{X-TC~\cite{DBLP:conf/cvpr/ZamirSCSCMG20} (IN$\times$2,\\enc$\times$1, dec$\times$2)} &\makecell[c]{CEN-dec (IN$\times$2,\\enc$\times$1, dec$\times$2)} &\makecell[c]{CEN-dec + X-TC~\cite{DBLP:conf/cvpr/ZamirSCSCMG20}\\(IN$\times$2, enc$\times$1, dec$\times$2)}\\
\midrule

\multirow{2}{*}{$ \text{RGB}\to\hskip-0.1em\left\{
\begin{aligned}
&\text{SemSeg} \\
&\text{Depth}\\
\end{aligned}
\right.$}&23.13 / 36.95&22.93 / 36.51 &20.19 / 35.09&19.44 / 34.36 & \textbf{18.79} / \textbf{33.75} \\
&4.93 / 8.75&4.72 / 8.50&4.17 / 7.60& 3.99 / 7.78 & \textbf{3.70} / \textbf{7.56} \\
\midrule
\multirow{2}{*}{$ \text{RGB}\to\hskip-0.1em\left\{
\begin{aligned}
&\text{Normal} \\
&\text{Curve} \\
\end{aligned}
\right.$}
&15.30 / 31.19 &15.92 / 32.04&14.35 / 30.28& 13.19 / 28.61& \textbf{12.02} / \textbf{27.66}  \\$\;\,$
& $\;\,$5.04 / 14.42 & $\;\,$4.97 / 14.22 &$\;\,$4.24 / 13.05& $\;\,$4.05 / 12.86& $\;\,$\textbf{3.32} / \textbf{11.29}\\
\midrule
Total params. (M)& Gen: 108.7; Dis: 8.3& Gen: 89.3; Dis: 8.3&Gen: 89.3; Dis: 8.3&Gen: 89.3; Dis: 8.3&Gen: 89.3; Dis: 8.3\\
\midrule
\midrule

\textbf{\makecell{Multimodal multitask\\learning}}&\makecell[c]{Indiv (IN$\times$4,\\enc$\times$2, dec$\times$2)}&\makecell[c]{CEN-enc (IN$\times$4,\\enc$\times$1, dec$\times$2)}&\makecell[c]{CEN-dec (IN$\times$4,\\enc$\times$1, dec$\times$2)} &\makecell[c]{CEN-enc \& dec\\(IN$\times$4, enc$\times$1, dec$\times$2)} \\
\midrule

\multirow{2}{*}{$\;\; \left.
\begin{aligned}
&\text{RGB} \\
&\text{Depth}\\
\end{aligned}
\right\}
\to\hskip-0.1em\left\{
\begin{aligned}
&\text{SemSeg} \\
&\text{Curve}\\
\end{aligned}
\right.$}&23.55 / 37.23&18.68 / 33.71&20.52 / 35.79&\textbf{17.86} / \textbf{32.94} \\
&$\;\,$4.94 / 14.22&$\;\,$4.25 / 13.30&$\;\,$4.63 / 13.87&$\;\,$\textbf{4.04} / \textbf{12.68}\\

\midrule
\multirow{2}{*}{$\;\, \left.
\begin{aligned}
&\text{RGB} \\
&\text{Depth}\\
\end{aligned}
\right\}
\to\hskip-0.1em\left\{
\begin{aligned}
&\text{Nomal} \\
&\text{Shade}\\
\end{aligned}
\right.$}&15.71 / 31.60&11.35 / 27.13&13.64 / 29.19&\textbf{10.62} / \textbf{25.28} \\
&$\;\,$7.09 / 11.43&$\;\,$6.13 / 9.97$\;$&$\;\,$6.94 / 10.65&$\;\,$\textbf{5.83} / \textbf{9.25}\\

\midrule
Total params. (M)& Gen: 108.7; Dis: 8.3& Gen: 89.3; Dis: 8.3&Gen: 89.3; Dis: 8.3&Gen: 89.3; Dis: 8.3\\

\bottomrule[1pt]
\end{tabular}}
\label{tabs:translation}
\vskip -0.03 in
\end{table*}

\textbf{Evaluation for  unsupervised learning.} Apart from the common supervised training settings for dense image prediction~\cite{wang2020cen,wang2022tokenfusion}, we show the potential of  CEN  for  unsupervised learning, \emph{e.g.},~\cite{DBLP:journals/inffus/MaML19, DBLP:journals/pami/XuMJGL22,DBLP:journals/corr/abs-2205-07179}.  As an example shown in Fig.~\ref{pic:silency} and Table~\ref{comUnSOD}, we  apply CEN to the Saliency Network in \cite{DBLP:journals/corr/abs-2205-07179} for RGB-Depth unsupervised saliency detection, which also achieves promising results, indicating the effectiveness of CEN in this case. Further discussion  of  unsupervised learning    is left for future exploration.

\textbf{Enlarging the sampled dataset.} 
In our image-to-image translation experiments on the Taskonomy dataset, we find that CEN with 1,000 training  images already achieves promising results. Here, we enlarge the sampled set with 15,000 training images and conduct the experiments. Since the training cost becomes quite large given the 15$\times$ expansion of the default sampled training dataset, we choose typical experiments to evaluate our CEN. Results provided in Table~\ref{tabs:translation} include multimodal fusion, cycle multimodal fusion, multitask learning, and multimodal multitask learning based on 15,000 sampled training images.  By comparison, we observe that mostly, training with 1,000 images and training with 15,000 images achieve  similar relative improvements of CEN over baselines. These results indicate that using 1,000 images for training already demonstrates the general advantages of CEN.



\bibliographystyle{IEEEtran}
\bibliography{IEEEabrv,bare_jrnl_compsoc}

\begin{thebibliography}{100}
\providecommand{\url}[1]{#1}
\csname url@samestyle\endcsname
\providecommand{\newblock}{\relax}
\providecommand{\bibinfo}[2]{#2}
\providecommand{\BIBentrySTDinterwordspacing}{\spaceskip=0pt\relax}
\providecommand{\BIBentryALTinterwordstretchfactor}{4}
\providecommand{\BIBentryALTinterwordspacing}{\spaceskip=\fontdimen2\font plus
\BIBentryALTinterwordstretchfactor\fontdimen3\font minus
  \fontdimen4\font\relax}
\providecommand{\BIBforeignlanguage}[2]{{%
\expandafter\ifx\csname l@#1\endcsname\relax
\typeout{** WARNING: IEEEtran.bst: No hyphenation pattern has been}%
\typeout{** loaded for the language `#1'. Using the pattern for}%
\typeout{** the default language instead.}%
\else
\language=\csname l@#1\endcsname
\fi
#2}}
\providecommand{\BIBdecl}{\relax}
\BIBdecl

\bibitem{journals/pami/BaltrusaitisAM19}
T.~Baltrusaitis, C.~Ahuja, and L.~Morency, ``Multimodal machine learning: {A}
  survey and taxonomy,'' \emph{{IEEE} Trans. PAMI}, 2019.

\bibitem{ramachandram2017deep}
D.~Ramachandram and G.~W. Taylor, ``Deep multimodal learning: A survey on
  recent advances and trends,'' \emph{{IEEE} Signal Processing Magazine}, 2017.

\bibitem{lin2019refinenet}
G.~Lin, F.~Liu, A.~Milan, C.~Shen, and I.~Reid, ``Refinenet: Multi-path
  refinement networks for dense prediction,'' \emph{{IEEE} Trans. PAMI}, 2019.

\bibitem{journals/ijcv/ValadaMB20}
A.~Valada, R.~Mohan, and W.~Burgard, ``Self-supervised model adaptation for
  multimodal semantic segmentation,'' \emph{IJCV}, 2020.

\bibitem{fan2018end}
L.~Fan, W.~Huang, C.~Gan, S.~Ermon, B.~Gong, and J.~Huang, ``End-to-end
  learning of motion representation for video understanding,'' in \emph{CVPR},
  2018.

\bibitem{conf/eccv/GarciaMM18}
N.~C. Garcia, P.~Morerio, and V.~Murino, ``Modality distillation with multiple
  stream networks for action recognition,'' in \emph{ECCV}, 2018.

\bibitem{journals/tip/SongLLG20}
S.~Song, J.~Liu, Y.~Li, and Z.~Guo, ``Modality compensation network:
  Cross-modal adaptation for action recognition,'' \emph{{IEEE} Trans. Image
  Process.}, 2020.

\bibitem{conf/iccv/AntolALMBZP15}
S.~Antol, A.~Agrawal, J.~Lu, M.~Mitchell, D.~Batra, C.~L. Zitnick, and
  D.~Parikh, ``{VQA:} visual question answering,'' in \emph{ICCV}, 2015.

\bibitem{conf/nips/IlievskiF17}
I.~Ilievski and J.~Feng, ``Multimodal learning and reasoning for visual
  question answering,'' in \emph{NIPS}, 2017.

\bibitem{conf/iccv/BalntasDSSKK17}
V.~Balntas, A.~Doumanoglou, C.~Sahin, J.~Sock, R.~Kouskouridas, and T.~Kim,
  ``Pose guided {RGBD} feature learning for 3d object pose estimation,'' in
  \emph{ICCV}, 2017.

\bibitem{conf/iclr/JinYBJ19}
W.~Jin, K.~Yang, R.~Barzilay, and T.~S. Jaakkola, ``Learning multimodal
  graph-to-graph translation for molecule optimization,'' in \emph{ICLR}, 2019.

\bibitem{conf/iccv/ZhangZSWSL19}
W.~Zhang, H.~Zhou, S.~Sun, Z.~Wang, J.~Shi, and C.~C. Loy, ``Robust
  multi-modality multi-object tracking,'' in \emph{ICCV}, 2019.

\bibitem{zhang2021survey}
Y.~Zhang and Q.~Yang, ``A survey on multi-task learning,'' \emph{{IEEE} Trans.
  Knowledge and Data Engineering}, 2021.

\bibitem{zhou2017multi}
D.~Zhou, J.~Wang, B.~Jiang, H.~Guo, and Y.~Li, ``Multi-task multi-view learning
  based on cooperative multi-objective optimization,'' \emph{IEEE Access},
  2017.

\bibitem{misra2016cross}
I.~Misra, A.~Shrivastava, A.~Gupta, and M.~Hebert, ``Cross-stitch networks for
  multi-task learning,'' in \emph{CVPR}, 2016.

\bibitem{liu2019end}
S.~Liu, E.~Johns, and A.~J. Davison, ``End-to-end multi-task learning with
  attention,'' in \emph{CVPR}, 2019.

\bibitem{guo2020learning}
P.~Guo, C.-Y. Lee, and D.~Ulbricht, ``Learning to branch for multi-task
  learning,'' in \emph{ICML}, 2020.

\bibitem{standley2020tasks}
T.~Standley, A.~Zamir, D.~Chen, L.~Guibas, J.~Malik, and S.~Savarese, ``Which
  tasks should be learned together in multi-task learning?'' in \emph{ICML},
  2020.

\bibitem{sun2020adashare}
X.~Sun, R.~Panda, R.~Feris, and K.~Saenko, ``Adashare: Learning what to share
  for efficient deep multi-task learning,'' \emph{NeurIPS}, 2020.

\bibitem{andreas2017modular}
J.~Andreas, D.~Klein, and S.~Levine, ``Modular multitask reinforcement learning
  with policy sketches,'' in \emph{ICML}, 2017.

\bibitem{rahmatizadeh2018vision}
R.~Rahmatizadeh, P.~Abolghasemi, L.~B{\"o}l{\"o}ni, and S.~Levine,
  ``Vision-based multi-task manipulation for inexpensive robots using
  end-to-end learning from demonstration,'' in \emph{ICRA}, 2018.

\bibitem{journals/corr/LongSD14}
J.~Long, E.~Shelhamer, and T.~Darrell, ``Fully convolutional networks for
  semantic segmentation,'' in \emph{CVPR}, 2015.

\bibitem{DBLP:journals/pami/ChenPKMY18}
L.~Chen, G.~Papandreou, I.~Kokkinos, K.~Murphy, and A.~L. Yuille, ``Deeplab:
  Semantic image segmentation with deep convolutional nets, atrous convolution,
  and fully connected crfs,'' \emph{{IEEE} Trans. PAMI}, 2018.

\bibitem{DBLP:conf/iccv/HuangLC021}
S.~Huang, Z.~Lu, R.~Cheng, and C.~He, ``Fapn: Feature-aligned pyramid network
  for dense image prediction,'' in \emph{ICCV}, 2021.

\bibitem{conf/cvpr/IsolaZZE17}
P.~Isola, J.~Zhu, T.~Zhou, and A.~A. Efros, ``Image-to-image translation with
  conditional adversarial networks,'' in \emph{CVPR}, 2017.

\bibitem{DBLP:conf/eccv/HuangLBK18}
X.~Huang, M.~Liu, S.~J. Belongie, and J.~Kautz, ``Multimodal unsupervised
  image-to-image translation,'' in \emph{ECCV}, 2018.

\bibitem{DBLP:journals/inffus/MaML19}
J.~Ma, Y.~Ma, and C.~Li, ``Infrared and visible image fusion methods and
  applications: {A} survey,'' \emph{Inf. Fusion}, 2019.

\bibitem{DBLP:journals/ijcv/LeeTMHLSY20}
H.~Lee, H.~Tseng, Q.~Mao, J.~Huang, Y.~Lu, M.~Singh, and M.~Yang, ``{DRIT++:}
  diverse image-to-image translation via disentangled representations,''
  \emph{IJCV}, 2020.

\bibitem{conf/accv/HazirbasMDC16}
C.~Hazirbas, L.~Ma, C.~Domokos, and D.~Cremers, ``Fusenet: Incorporating depth
  into semantic segmentation via fusion-based {CNN} architecture,'' in
  \emph{ACCV}, 2016.

\bibitem{conf/cvpr/ZengTHYSCW19}
J.~Zeng, Y.~Tong, Y.~Huang, Q.~Yan, W.~Sun, J.~Chen, and Y.~Wang, ``Deep
  surface normal estimation with hierarchical {RGB-D} fusion,'' in \emph{CVPR},
  2019.

\bibitem{conf/eccv/WangWTSW16}
J.~Wang, Z.~Wang, D.~Tao, S.~See, and G.~Wang, ``Learning common and specific
  features for {RGB-D} semantic segmentation with deconvolutional networks,''
  in \emph{ECCV}, 2016.

\bibitem{kokkinos2017ubernet}
I.~Kokkinos, ``Ubernet: Training a universal convolutional neural network for
  low-, mid-, and high-level vision using diverse datasets and limited
  memory,'' in \emph{CVPR}, 2017.

\bibitem{chennupati2019multinet++}
S.~Chennupati, G.~Sistu, S.~Yogamani, and S.~A~Rawashdeh, ``Multinet++:
  Multi-stream feature aggregation and geometric loss strategy for multi-task
  learning,'' in \emph{CVPR Workshops}, 2019.

\bibitem{ruder2019latent}
S.~Ruder, J.~Bingel, I.~Augenstein, and A.~S{\o}gaard, ``Latent multi-task
  architecture learning,'' in \emph{AAAI}, 2019.

\bibitem{conf/cvpr/DuWWZW19}
D.~Du, L.~Wang, H.~Wang, K.~Zhao, and G.~Wu, ``Translate-to-recognize networks
  for {RGB-D} scene recognition,'' in \emph{CVPR}, 2019.

\bibitem{conf/iccv/LeePH17}
S.~Lee, S.~Park, and K.~Hong, ``Rdfnet: {RGB-D} multi-level residual feature
  fusion for indoor semantic segmentation,'' in \emph{ICCV}, 2017.

\bibitem{sener2018multi}
O.~Sener and V.~Koltun, ``Multi-task learning as multi-objective
  optimization,'' \emph{arXiv preprint arXiv:1810.04650}, 2018.

\bibitem{conf/iccv/LiuLSHYZ17}
Z.~Liu, J.~Li, Z.~Shen, G.~Huang, S.~Yan, and C.~Zhang, ``Learning efficient
  convolutional networks through network slimming,'' in \emph{ICCV}, 2017.

\bibitem{conf/iclr/YeL0W18}
J.~Ye, X.~Lu, Z.~Lin, and J.~Z. Wang, ``Rethinking the
  smaller-norm-less-informative assumption in channel pruning of convolution
  layers,'' in \emph{ICLR}, 2018.

\bibitem{conf/icml/IoffeS15}
S.~Ioffe and C.~Szegedy, ``Batch normalization: Accelerating deep network
  training by reducing internal covariate shift,'' in \emph{ICML}, 2015.

\bibitem{DBLP:journals/corr/UlyanovVL16}
D.~Ulyanov, A.~Vedaldi, and V.~S. Lempitsky, ``Instance normalization: The
  missing ingredient for fast stylization,'' \emph{arXiv preprint
  arXiv:1607.08022}, 2016.

\bibitem{conf/eccv/SilbermanHKF12}
N.~Silberman, D.~Hoiem, P.~Kohli, and R.~Fergus, ``Indoor segmentation and
  support inference from {RGBD} images,'' in \emph{ECCV}, 2012.

\bibitem{conf/cvpr/SongLX15}
S.~Song, S.~P. Lichtenberg, and J.~Xiao, ``{SUN} {RGB-D:} {A} {RGB-D} scene
  understanding benchmark suite,'' in \emph{CVPR}, 2015.

\bibitem{conf/cvpr/ZamirSSGMS18}
A.~R. Zamir, A.~Sax, W.~B. Shen, L.~J. Guibas, J.~Malik, and S.~Savarese,
  ``Taskonomy: Disentangling task transfer learning,'' in \emph{CVPR}, 2018.

\bibitem{conf/icml/NgiamKKNLN11}
J.~Ngiam, A.~Khosla, M.~Kim, J.~Nam, H.~Lee, and A.~Y. Ng, ``Multimodal deep
  learning,'' in \emph{ICML}, 2011.

\bibitem{DBLP:journals/expert/ZhouYLL21}
W.~Zhou, J.~Yuan, J.~Lei, and T.~Luo, ``Tsnet: Three-stream self-attention
  network for {RGB-D} indoor semantic segmentation,'' \emph{{IEEE} Intell.
  Syst.}, 2021.

\bibitem{DBLP:journals/pami/XuMJGL22}
H.~Xu, J.~Ma, J.~Jiang, X.~Guo, and H.~Ling, ``U2fusion: {A} unified
  unsupervised image fusion network,'' \emph{{IEEE} Trans. PAMI}, 2022.

\bibitem{conf/iccv/LinCCHH17}
D.~Lin, G.~Chen, D.~Cohen{-}Or, P.~Heng, and H.~Huang, ``Cascaded feature
  network for semantic segmentation of {RGB-D} images,'' in \emph{ICCV}, 2017.

\bibitem{journals/jmlr/GrettonBRSS12}
A.~Gretton, K.~M. Borgwardt, M.~J. Rasch, B.~Sch{\"{o}}lkopf, and A.~J. Smola,
  ``A kernel two-sample test,'' in \emph{JMLR}, 2012.

\bibitem{conf/nips/BousmalisTSKE16}
K.~Bousmalis, G.~Trigeorgis, N.~Silberman, D.~Krishnan, and D.~Erhan, ``Domain
  separation networks,'' in \emph{NIPS}, 2016.

\bibitem{atrey2010multimodal}
P.~K. Atrey, M.~A. Hossain, A.~El~Saddik, and M.~S. Kankanhalli, ``Multimodal
  fusion for multimedia analysis: a survey,'' \emph{Multimedia systems}, 2010.

\bibitem{bruni2014multimodal}
E.~Bruni, N.-K. Tran, and M.~Baroni, ``Multimodal distributional semantics,''
  in \emph{Journal of Artificial Intelligence Research}, 2014.

\bibitem{hall1997introduction}
D.~L. Hall and J.~Llinas, ``An introduction to multisensor data fusion,''
  \emph{Proceedings of the IEEE}, 1997.

\bibitem{snoek2005early}
C.~G. Snoek, M.~Worring, and A.~W. Smeulders, ``Early versus late fusion in
  semantic video analysis,'' in \emph{ACM MM}, 2005.

\bibitem{lazaridou2014wampimuk}
A.~Lazaridou, E.~Bruni, and M.~Baroni, ``Is this a wampimuk? cross-modal
  mapping between distributional semantics and the visual world,'' in
  \emph{ACL}, 2014.

\bibitem{wang2020asymfusion}
Y.~Wang, F.~Sun, M.~Lu, and A.~Yao, ``Learning deep multimodal feature
  representation with asymmetric multi-layer fusion,'' in \emph{ACM MM}, 2020.

\bibitem{conf/nips/VriesSMLPC17}
H.~de~Vries, F.~Strub, J.~Mary, H.~Larochelle, O.~Pietquin, and A.~C.
  Courville, ``Modulating early visual processing by language,'' in
  \emph{NIPS}, 2017.

\bibitem{de2017guesswhat}
H.~De~Vries, F.~Strub, S.~Chandar, O.~Pietquin, H.~Larochelle, and
  A.~Courville, ``Guesswhat?! visual object discovery through multi-modal
  dialogue,'' in \emph{CVPR}, 2017.

\bibitem{dumoulin2018feature}
V.~Dumoulin, E.~Perez, N.~Schucher, F.~Strub, H.~d. Vries, A.~Courville, and
  Y.~Bengio, ``Feature-wise transformations,'' in \emph{Distill}, 2018.

\bibitem{long2015learning}
M.~Long, Z.~Cao, J.~Wang, and P.~S. Yu, ``Learning multiple tasks with
  multilinear relationship networks,'' \emph{arXiv preprint arXiv:1506.02117},
  2015.

\bibitem{suteu2019regularizing}
M.~Suteu and Y.~Guo, ``Regularizing deep multi-task networks using orthogonal
  gradients,'' \emph{arXiv preprint arXiv:1912.06844}, 2019.

\bibitem{gao2019nddr}
Y.~Gao, J.~Ma, M.~Zhao, W.~Liu, and A.~L. Yuille, ``Nddr-cnn: Layerwise feature
  fusing in multi-task cnns by neural discriminative dimensionality
  reduction,'' in \emph{CVPR}, 2019.

\bibitem{DBLP:journals/corr/ZhangY17aa}
Y.~Zhang and Q.~Yang, ``A survey on multi-task learning,'' \emph{arXiv preprint
  arXiv:1707.08114}, 2017.

\bibitem{DBLP:conf/icml/StandleyZCGMS20}
T.~Standley, A.~R. Zamir, D.~Chen, L.~J. Guibas, J.~Malik, and S.~Savarese,
  ``Which tasks should be learned together in multi-task learning?'' in
  \emph{ICML}, 2020.

\bibitem{DBLP:conf/cvpr/ZamirSCSCMG20}
A.~R. Zamir, A.~Sax, N.~Cheerla, R.~Suri, Z.~Cao, J.~Malik, and L.~J. Guibas,
  ``Robust learning through cross-task consistency,'' in \emph{CVPR}, 2020.

\bibitem{DBLP:conf/eccv/VandenhendeGG20}
S.~Vandenhende, S.~Georgoulis, and L.~V. Gool, ``Mti-net: Multi-scale task
  interaction networks for multi-task learning,'' in \emph{ECCV}, 2020.

\bibitem{DBLP:conf/iccv/ZhuPIE17}
J.~Zhu, T.~Park, P.~Isola, and A.~A. Efros, ``Unpaired image-to-image
  translation using cycle-consistent adversarial networks,'' in \emph{ICCV},
  2017.

\bibitem{shao2020channel}
W.~Shao, S.~Tang, X.~Pan, P.~Tan, X.~Wang, and P.~Luo, ``Channel equilibrium
  networks for learning deep representation,'' in \emph{ICML}, 2020.

\bibitem{conf/cvpr/ZhangZLS18}
X.~Zhang, X.~Zhou, M.~Lin, and J.~Sun, ``Shufflenet: An extremely efficient
  convolutional neural network for mobile devices,'' in \emph{CVPR}, 2018.

\bibitem{conf/cvpr/ChangYSKH19}
W.~Chang, T.~You, S.~Seo, S.~Kwak, and B.~Han, ``Domain-specific batch
  normalization for unsupervised domain adaptation,'' in \emph{CVPR}, 2019.

\bibitem{DBLP:conf/cvpr/Kokkinos17}
I.~Kokkinos, ``Ubernet: Training a universal convolutional neural network for
  low-, mid-, and high-level vision using diverse datasets and limited
  memory,'' in \emph{CVPR}, 2017.

\bibitem{conf/cvpr/GuptaAM13}
S.~Gupta, P.~Arbelaez, and J.~Malik, ``Perceptual organization and recognition
  of indoor scenes from {RGB-D} images,'' in \emph{CVPR}, 2013.

\bibitem{conf/cvpr/ZhaoSQWJ17}
H.~Zhao, J.~Shi, X.~Qi, X.~Wang, and J.~Jia, ``Pyramid scene parsing network,''
  in \emph{CVPR}, 2017.

\bibitem{conf/cvpr/HeZRS16}
K.~He, X.~Zhang, S.~Ren, and J.~Sun, ``Deep residual learning for image
  recognition,'' in \emph{CVPR}, 2016.

\bibitem{journals/ijcv/RussakovskyDSKS15}
O.~Russakovsky, J.~Deng, H.~Su, J.~Krause, S.~Satheesh, S.~Ma, Z.~Huang,
  A.~Karpathy, A.~Khosla, M.~S. Bernstein, A.~C. Berg, and F.~Li, ``Imagenet
  large scale visual recognition challenge,'' \emph{IJCV}, 2015.

\bibitem{conf/nips/HeuselRUNH17}
M.~Heusel, H.~Ramsauer, T.~Unterthiner, B.~Nessler, and S.~Hochreiter, ``Gans
  trained by a two time-scale update rule converge to a local nash
  equilibrium,'' in \emph{NIPS}, 2017.

\bibitem{conf/iclr/BinkowskiSAG18}
M.~Binkowski, D.~J. Sutherland, M.~Arbel, and A.~Gretton, ``Demystifying {MMD}
  gans,'' in \emph{ICLR}, 2018.

\bibitem{conf/icip/HuYFW19}
X.~Hu, K.~Yang, L.~Fei, and K.~Wang, ``{ACNET:} attention based network to
  exploit complementary features for {RGBD} semantic segmentation,'' in
  \emph{ICIP}, 2019.

\bibitem{conf/iccv/QiLJFU17}
X.~Qi, R.~Liao, J.~Jia, S.~Fidler, and R.~Urtasun, ``3d graph neural networks
  for {RGBD} semantic segmentation,'' in \emph{ICCV}, 2017.

\bibitem{journals/tcyb/LinZJLH20}
D.~Lin, R.~Zhang, Y.~Ji, P.~Li, and H.~Huang, ``{SCN:} switchable context
  network for semantic segmentation of {RGB-D} images,'' \emph{{IEEE} Trans.
  Cybern.}, 2020.

\bibitem{DBLP:conf/nips/SunPFS20}
X.~Sun, R.~Panda, R.~Feris, and K.~Saenko, ``Adashare: Learning what to share
  for efficient deep multi-task learning,'' in \emph{NeurIPS}, 2020.

\bibitem{DBLP:journals/corr/abs-2205-07179}
W.~Ji, J.~Li, Q.~Bi, C.~Guo, J.~Liu, and L.~Cheng, ``Promoting saliency from
  depth: Deep unsupervised {RGB-D} saliency detection,'' in \emph{ICLR}, 2022.

\bibitem{conf/cvpr/CordtsORREBFRS16}
M.~Cordts, M.~Omran, S.~Ramos, T.~Rehfeld, M.~Enzweiler, R.~Benenson,
  U.~Franke, S.~Roth, and B.~Schiele, ``The cityscapes dataset for semantic
  urban scene understanding,'' in \emph{CVPR}, 2016.

\bibitem{NJU2K}
R.~Ju, L.~Ge, W.~Geng, T.~Ren, and G.~Wu, ``Depth saliency based on anisotropic
  center-surround difference,'' in \emph{ICIP}, 2014.

\bibitem{NLPR}
H.~Peng, B.~Li, W.~Xiong, W.~Hu, and R.~Ji, ``{{RGB}D} salient object
  detection: a benchmark and algorithms,'' in \emph{ECCV}, 2014.

\bibitem{STERE}
Y.~Niu, Y.~Geng, X.~Li, and F.~Liu, ``Leveraging stereopsis for saliency
  analysis,'' in \emph{CVPR}, 2012.

\bibitem{DMRA2019}
Y.~Piao, W.~Ji, J.~Li, M.~Zhang, and H.~Lu, ``Depth-induced multi-scale
  recurrent attention network for saliency detection,'' in \emph{ICCV}, 2019.

\bibitem{borji2015salient}
A.~Borji, M.-M. Cheng, H.~Jiang, and J.~Li, ``Salient object detection: A
  benchmark,'' \emph{{IEEE} Trans. Image Process.}, 2015.

\bibitem{MST2016}
W.-C. Tu, S.~He, Q.~Yang, and S.-Y. Chien, ``Real-time salient object detection
  with a minimum spanning tree,'' in \emph{CVPR}, 2016.

\bibitem{BSCA2015}
Y.~Qin, H.~Lu, Y.~Xu, and H.~Wang, ``Saliency detection via cellular
  automata,'' in \emph{CVPR}, 2015.

\bibitem{GP2015}
J.~Ren, X.~Gong, L.~Yu, W.~Zhou, and M.~Ying~Yang, ``Exploiting global priors
  for {RGB-D} saliency detection,'' in \emph{CVPR Workshops}, 2015.

\bibitem{CDB2018}
F.~Liang, L.~Duan, W.~Ma, Y.~Qiao, Z.~Cai, and L.~Qing, ``Stereoscopic saliency
  model using contrast and depth-guided-background prior,''
  \emph{Neurocomputing}, vol. 275, 2018.

\bibitem{SE2016}
J.~Guo, T.~Ren, and J.~Bei, ``Salient object detection for {RGB-D} image via
  saliency evolution,'' in \emph{ICME}, 2016.

\bibitem{DCMC}
R.~Cong, J.~Lei, C.~Zhang, Q.~Huang, X.~Cao, and C.~Hou, ``Saliency detection
  for stereoscopic images based on depth confidence analysis and multiple cues
  fusion,'' \emph{{IEEE} Signal Processing Letters}, 2016.

\bibitem{MB2017}
C.~Zhu, G.~Li, X.~Guo, W.~Wang, and R.~Wang, ``A multilayer backpropagation
  saliency detection algorithm based on depth mining,'' in \emph{ICCAIP}, 2017.

\bibitem{CDCP2017}
C.~Zhu, G.~Li, W.~Wang, and R.~Wang, ``An innovative salient object detection
  using center-dark channel prior,'' in \emph{ICCV Workshops}, 2017.

\bibitem{USD2018}
J.~Zhang, T.~Zhang, Y.~Dai, M.~Harandi, and R.~Hartley, ``Deep unsupervised
  saliency detection: A multiple noisy labeling perspective,'' in \emph{CVPR},
  2018.

\bibitem{DeepUSPS2019}
T.~Nguyen, M.~Dax, C.~K. Mummadi, N.~Ngo, T.~H.~P. Nguyen, Z.~Lou, and T.~Brox,
  ``{DeepUSPS}: Deep robust unsupervised saliency prediction via
  self-supervision,'' in \emph{NeurIPS}, 2019.

\bibitem{wang2020cen}
Y.~Wang, W.~Huang, F.~Sun, T.~Xu, Y.~Rong, and J.~Huang, ``Deep multimodal
  fusion by channel exchanging,'' in \emph{NeurIPS}, 2020.

\bibitem{wang2022tokenfusion}
Y.~Wang, X.~Chen, L.~Cao, W.~Huang, F.~Sun, and Y.~Wang, ``Multimodal token
  fusion for vision transformers,'' in \emph{CVPR}, 2022.

\end{thebibliography}

\end{document}